\documentclass[a4paper,fleqn]{cas-sc}

\usepackage{caption}
\usepackage[normalem]{ulem}
\usepackage{subcaption}

\usepackage[authoryear,longnamesfirst]{natbib}

\def\tsc#1{\csdef{#1}{\textsc{\lowercase{#1}}\xspace}}
\tsc{WGM}
\tsc{QE}
\begin{document}

\let\WriteBookmarks\relax
\def\floatpagepagefraction{1}
\def\textpagefraction{.001}

\shorttitle{}    

\shortauthors{Z. Chu et~al.}  

\title [mode = title]{Multi-Relational Knowledge Graph Enhanced Embedding for Trajectory-User Linking}  

\tnotemark[1] 

\tnotetext[1]{} 


\author[1]{Zhifeng Chu}%[<options>]


\ead{1000569132@smail.shnu.edu.cn}

\ead[url]{}

\credit{}

%Address/affiliation
 \affiliation[1]{organization={the College of Information, Mechanical
and Electrical Engineering, Shanghai Normal University},
            %addressline={100,Guilin Road}, 
            city={Shanghai},
% %          citysep={}, % Uncomment if no comma needed between city and postcode
             postcode={201400}, 
             country={China}}

\author[1]{Bin Wang}%[]
\cormark[1]

% Footnote of the second author

% Email id of the second author
\ead{binwang@shnu.edu.cn}

\ead[url]{}

\credit{}

 
\author[2]{Xi Zhai}%[]

% Footnote of the second author
%\fnmark[3]

% Email id of the second author
\ead{jessie_zx28@163.com}

\ead[url]{}

\credit{}

 \affiliation[2]{organization={Shanghai Urban-Rural Construction \& Transportation Development Institute},
            city={Shanghai},
% %          citysep={}, % Uncomment if no comma needed between city and postcode
             postcode={200032}, 
             country={China}}

\author[1]{Yan Ma}%[]

% Footnote of the second author

% Email id of the second author
\ead{ma-yan@shnu.edu.cn}

\ead[url]{}

\credit{}

% \affiliation[2]{organization={},
% %          citysep={}, % Uncomment if no comma needed between city and postcode

\author[3]{Xiong Li}
\cormark[1]
\ead[url]{}
\cortext[1]{Corresponding author}
\affiliation[3]{organization={Zhongke Zidong Information Technology},
            city={Beijing},
% %          citysep={}, % Uncomment if no comma needed between city and postcode
             postcode={100000}, 
             country={China}}
\ead{li.xiong@foxmail.com}
\credit{}

\fntext[1]{}


\begin{abstract}
Trajectory-User Linking (TUL) aims to identify the owner of an anonymous trajectory from a set of candidate users, providing a basis for user mobility analysis and personalized location-aware services. Existing methods often learn Point of Interest (POI), temporal, and semantic features independently, make limited use of structural knowledge shared across trajectories, and compress structural and sequential information before classification. To address these issues, we propose \textbf{M}ulti-Rel\textbf{a}tional \textbf{K}nowledge Graph
\textbf{E}nhanced Embedding for
\textbf{T}rajectory-\textbf{U}ser \textbf{L}inking (MakeTUL), which, to the best of our knowledge, is the first attempt to introduce knowledge graph representation learning into TUL. MakeTUL organizes visit-time, POI-category, and transfer-speed information as typed relations in a multi-relational mobility knowledge graph, allowing heterogeneous mobility semantics to jointly constrain the learned embeddings. The resulting POI representations are further enriched with high-order co-occurrence patterns extracted from the trajectory collection, providing structural prior knowledge for sparse and overlapping trajectories. By integrating these prior-enhanced representations with temporal, category, and transfer information, the trajectory sequence learning module captures ordered mobility patterns, while a dual-branch classification layer preserves and combines global structural evidence and sequential evidence at the decision level. Experiments on Foursquare-NYC, Foursquare-TKY, and Foursquare-JKT are evaluated using ACC@1, ACC@5, Macro-Precision, Macro-Recall, and Macro-F1. MakeTUL achieves the best results on all five metrics across six user-scale settings. On Foursquare-JKT with 1,400 users, ACC@1 and Macro-F1 exceed the strongest baselines by 10.67\% and 12.70\%, respectively, showing that multi-relational knowledge and structural priors provide useful complementary information for trajectory owner classification, particularly under larger candidate-user settings.The source code is available at \url{https://github.com/superior-DL/MakeTUL}.
\end{abstract}


% \begin{highlights}
% \item 
% \item 
% \item 
% \end{highlights}

% Keywords

\maketitle

% Main text
% Replace each \refneeded{...} with the corresponding citation later.
\newcommand{\refneeded}[1]{\textnormal{[REFERENCE NEEDED: #1]}}
\newcommand{\resultneeded}{\textnormal{[RESULT TO BE ADDED]}}
\newcommand{\datasetneeded}{\textnormal{[DATASET STATISTICS NEEDED]}}

\section{Introduction}

The widespread deployment of Global Positioning System (GPS)-enabled
devices and Location-Based Social Networks (LBSNs) has generated large
volumes of check-in trajectories that record users' visits to different
locations over time \cite{zheng2015trajectory,graser2024mobilitydl,kong2023mobility}. These trajectories provide an important data source for
understanding user mobility patterns and supporting location-aware services \cite{yabe2024yjmob100k,du2025review}.
TULER \cite{gao2017identifying} first formulated and defined Trajectory-User Linking (TUL) as the
task of associating an anonymous trajectory with the user who most likely
generated it. Given a set of anonymous trajectories and a set of candidate
users, TUL learns a mapping from each anonymous trajectory to its corresponding
trajectory owner. Unlike conventional trajectory classification, which usually
distinguishes a limited number of transportation modes or activity types, TUL
treats candidate users as classification categories. The resulting large label
space, together with the sparsity and irregularity of check-in records, makes
the extraction of discriminative user mobility patterns particularly
challenging.

Before deep representation learning became the dominant solution, TUL and
related trajectory classification tasks were mainly addressed using trajectory
similarity measures and statistical learning models. Longest Common
Sub-Sequence (LCSS) \cite{ying2010mining} measures the similarity between two trajectories by
identifying their longest matched subsequence. Dynamic Time Warping (DTW) \cite{keogh2000scaling}
aligns trajectories with different temporal progressions and calculates their
alignment cost. Spatial-Temporal Similarity (STS) \cite{li2021spatial} evaluates the consistency
of trajectories by jointly considering spatial and temporal proximity.
Signature Representation (SR) \cite{jin2019moving} constructs compact trajectory signatures for
trajectory comparison and retrieval. Linear Discriminant Analysis (LDA) \cite{shahdoosti2017spectral}
classifies trajectories by projecting manually constructed features into a
discriminative subspace, whereas Support Vector Machine (SVM) \cite{cortes1995support} learns a
decision boundary between user classes from trajectory feature vectors.
Decision Tree (DT) \cite{jiang2018survey} and Random Forest (RF) \cite{breiman2001random} have also been applied to
mobility pattern classification using statistical trajectory features. Although
these methods provide interpretable solutions, their performance depends
strongly on predefined distance functions and feature engineering, which limits
their ability to represent sparse, irregular, and high-dimensional check-in
trajectories.

Deep learning methods subsequently shifted the focus from manually designed
features to trajectory representation learning. TULER \cite{gao2017identifying} learns POI embeddings
using a Word2Vec-based strategy and employs recurrent neural networks (RNNs) to
capture sequential POI transition patterns. TULVAE \cite{zhou2018trajectory} introduces a
semi-supervised variational autoencoder to exploit unlabeled trajectories and
model the hierarchical latent semantics of human mobility. DeepTUL \cite{miao2020trajectory} combines
historical trajectories with an attentive recurrent architecture to capture
multi-periodic user mobility patterns. GNNTUL \cite{zhou2021trajectory} constructs a check-in graph
that integrates geographical proximity and POI transition information and
applies a graph neural network to learn location representations. MainTUL \cite{chen2022mutual}
uses an input-trajectory branch and a historical-trajectory branch to transfer
mobility knowledge through mutual distillation. S2TUL \cite{deng2023s2tul} constructs multiple
trajectory-level homogeneous and heterogeneous graphs to model repeatability,
spatial proximity, and spatio-temporal proximity among trajectories. AttnTUL \cite{chen2024trajectory}
introduces hierarchical spatio-temporal attention to jointly model local
transition dependencies and global trajectory relationships. TULMGAT \cite{li2025trajectory}
constructs multi-scale check-in-oriented graphs and learns trajectory
representations through graph attention and cross-scale information fusion.
ScaleTUL \cite{zhang2025scalable} combines spatio-temporal trajectory augmentation with a dual-stream
representation network to capture short-term and long-term mobility
dependencies in large-scale TUL settings. HGTUL \cite{10.1007/978-981-95-7075-1_21} models POIs as vertices and
trajectories as hyperedges to capture high-order inter-trajectory associations
induced by shared POIs. These studies have progressively strengthened
sequential, spatial, and structural trajectory modeling, but the embedding,
trajectory encoding, and classification stages still leave several aspects
insufficiently explored.

First, existing embedding schemes usually rely on randomly initialized lookup
tables or Word2Vec-based POI pretraining, without explicitly modeling the
relations between POIs and other mobility semantics. As shown in
Fig.~\ref{fig:semantic_embedding}(a), the Continuous Bag-of-Words (CBOW) \cite{mikolov2013distributed}
objective uses the neighboring POIs $p_1$ and $p_3$ to predict the center POI
$p_2$, thereby learning POI representations from local co-occurrence contexts.
However, visit time, POI category, and transition semantics are generally
represented by independently initialized embeddings and subsequently combined
with POI embeddings through addition or concatenation. Such operations treat
different semantic factors as separate identifiers rather than relationally
constrained knowledge. Figure~\ref{fig:semantic_embedding}(b) provides a more
structured description of the same trajectory. For example, $p_1$ and $p_2$
belong to category $c_1$ and are visited at time entities $t_1$ and $t_2$,
respectively; $p_3$ and $p_4$ belong to category $c_2$ and correspond to
$t_3$ and $t_4$; and $p_5$ is associated with category $c_3$ and time entity
$t_5$. Meanwhile, the directed transitions among consecutive POIs preserve
the movement order and provide the basis for distinguishing different
transfer-speed relations. Existing independent embedding strategies cannot
fully preserve these heterogeneous associations. Second, trajectory encoding
is commonly driven by the observed POI sequence of the current trajectory,
with limited incorporation of prior structural knowledge extracted from the
trajectory collection. Sequence-based methods mainly describe local transition
dependencies, whereas graph-based structural information is often processed
independently and fused only after trajectory encoding. Consequently, the POI
representations entering the sequence learning process are not directly
conditioned on high-order POI co-occurrence patterns across trajectories, which
is particularly restrictive when an anonymous trajectory is sparse or shares
many POIs with different users. Third, many existing methods combine structural
and sequential representations through addition or concatenation and then use
a single classification branch. This early compression does not retain the
independent prediction evidence provided by global trajectory structure and
fine-grained sequential patterns, making it difficult to adjust their relative
confidence at the decision level.

\begin{center}
\centering
    \includegraphics[width=0.8\linewidth,trim=0.2cm 0.2cm 0.2cm 0.2cm, clip]{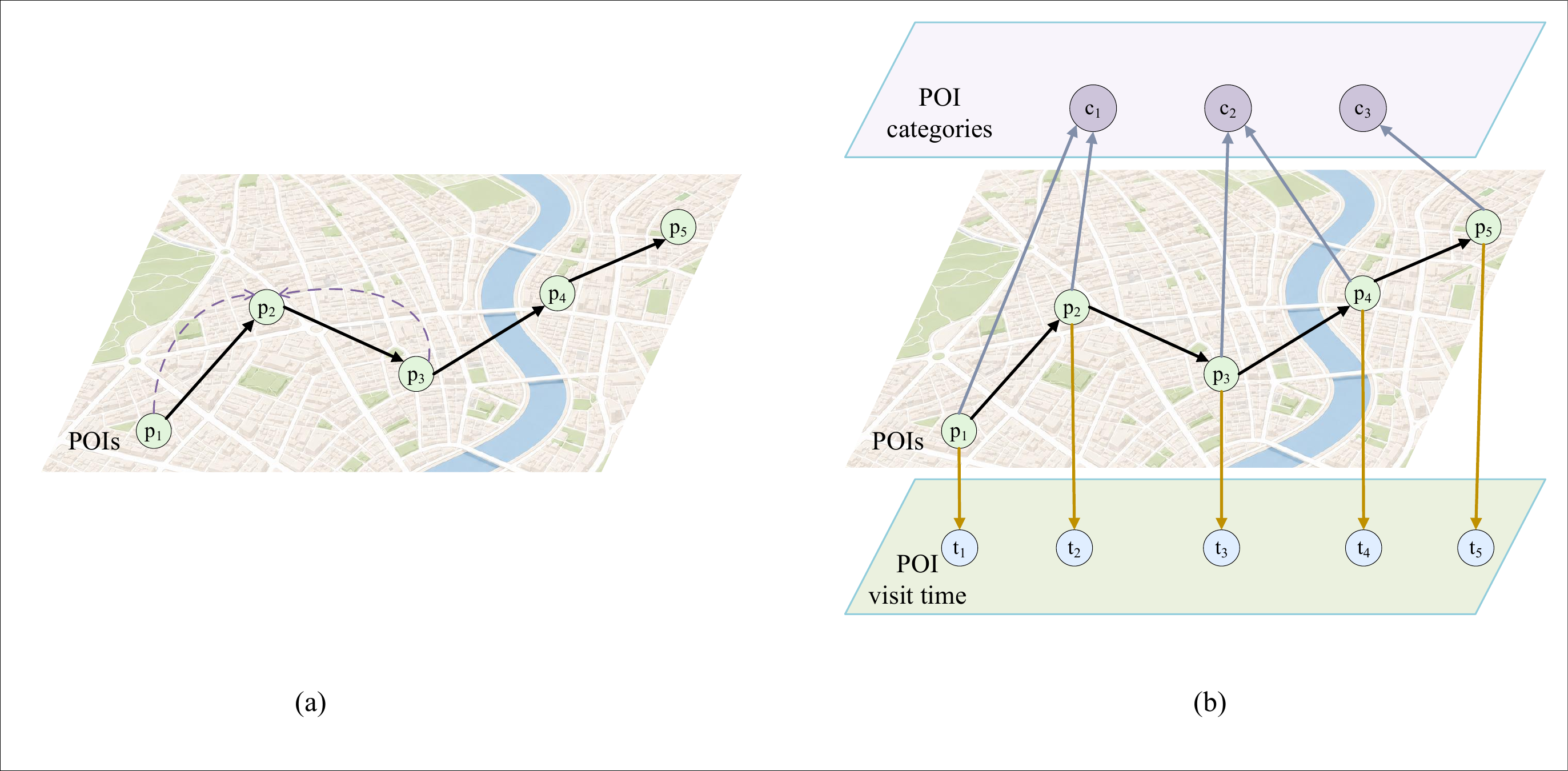}
    \captionof{figure}{Comparison between context-based POI embedding and
    multi-relational semantic association modeling. (a) CBOW learns the
    representation of $p_2$ from its neighboring POIs $p_1$ and $p_3$.
    (b) POIs are connected to visit-time and category entities through typed
    associations, while directed POI transitions are retained for
    relation-aware mobility modeling.}
    \label{fig:semantic_embedding}
\end{center}

To address these issues, we propose \emph{\textbf{M}ulti-Rel\textbf{a}tional \textbf{K}nowledge Graph
\textbf{E}nhanced Embedding for \textbf{T}rajectory-\textbf{U}ser \textbf{L}inking} (MakeTUL). To the best of
our knowledge, this is the first work to introduce knowledge graph
representation learning into the TUL task. MakeTUL constructs a
multi-relational mobility knowledge graph containing visit-time relations,
POI-category relations, and transfer-speed relations, and projects entities
into relation-specific spaces. Through multimodal embedding, the resulting
POI, time, category, and transfer-relation embeddings are jointly constrained
by the multi-relational structure rather than learned as mutually independent
lookup vectors. Based on the knowledge graph enhanced POI embeddings, the
\emph{Trajectory Prior Structure Extraction Module} further incorporates prior
structural knowledge into POI representations. These representations, together
with time, category, and transfer-relation embeddings, are then passed to the
\emph{Trajectory Sequence Learning Module} to model the ordered interactions
among prior-enhanced check-in representations. Therefore, the sequence
representation of an anonymous trajectory depends not only on its observed POI
order but also on global structural regularities extracted from the trajectory
collection. When the current trajectory is sparse or its visited POIs overlap
with those of multiple users, the extracted prior knowledge provides additional
structural context for trajectory encoding. At the prediction stage, MakeTUL
retains the global trajectory representation and the sequence representation as
two independent prediction branches in the \emph{Dual-Branch Classification
Layer}. Each branch produces its own user prediction scores, after which a
learnable temperature-calibrated fusion mechanism combines the two prediction
distributions. This design preserves the different evidence conveyed by
trajectory prior structure and sequential mobility patterns and adjusts the
confidence of the two branches at the prediction level. The TUL classification
objective and the knowledge graph representation objective are jointly optimized
so that multi-relational embedding learning directly supports trajectory owner
prediction.

The main contributions of this work are summarized as follows:

\begin{itemize}
    \item We propose a multi-relational knowledge graph enhanced multimodal
    embedding method for TUL. Visit-time, POI-category, and transfer-speed
    semantics are organized as typed relations, enabling POI, time, category,
    and transfer-relation embeddings to be jointly constrained in
    relation-specific representation spaces.

    \item We develop a prior-enhanced trajectory representation architecture
    consisting of the \emph{Trajectory Prior Structure Extraction Module} and
    the \emph{Trajectory Sequence Learning Module}. The former extracts prior
    structural knowledge from high-order POI co-occurrence patterns, while the
    latter integrates this knowledge with multimodal semantic embeddings to
    model the ordered mobility patterns of anonymous trajectories.

    \item We design a \emph{Dual-Branch Classification Layer} that preserves
    the global trajectory representation and the sequence representation in
    separate prediction branches and combines their user prediction
    distributions through learnable temperature-calibrated fusion.

    \item We conduct extensive experiments on three real-world datasets.
    Comparisons with representative TUL methods and component-wise analyses
    validate the effectiveness of MakeTUL.
\end{itemize}

\section{Related work}

Existing deep learning methods for TUL can be
broadly divided into two categories according to how they organize mobility
information: sequence-based modeling methods and graph-based modeling
methods. Sequence-based methods represent an anonymous trajectory as an
ordered sequence of check-ins and focus on capturing intra-trajectory
dependencies. Graph-based methods organize POIs or trajectories as graph
elements and exploit spatial or inter-trajectory structures.
The following subsections review these two research directions.

\subsection{Sequence-based modeling methods}

Sequence-based methods regard a trajectory as a chronologically ordered POI
sequence and learn a trajectory representation by modeling successive check-in
records. TULER \cite{gao2017identifying} first formulates the TUL problem and employs a
Word2Vec-based embedding strategy to map POIs into a low-dimensional space.
The resulting POI embeddings are processed by RNNs to
capture sequential POI transitions and predict the trajectory owner. Although
TULER establishes the basic sequence-classification paradigm for TUL, its
supervised learning process is sensitive to sparse labeled trajectories.
TULVAE \cite{zhou2018trajectory} extends recurrent trajectory modeling with a semi-supervised
Variational Autoencoder (VAE). It introduces latent variables at different
levels of the recurrent architecture to model hierarchical mobility semantics
and incorporates unlabeled trajectories into model training. STULIG \cite{zhou2020toward} also
adopts a generative learning framework. It pretrains POI representations with
the Continuous Bag-of-Words technique and learns disentangled trajectory
representations from labeled and unlabeled mobility data. These generative
methods alleviate the dependence on labeled samples, but their trajectory
representations remain primarily derived from the sequential organization of
individual trajectories.
DeepTUL \cite{miao2020trajectory} introduces an attentive recurrent architecture to exploit historical
trajectories. It models periodic mobility regularities at different temporal
scales and uses attention to identify historical movement patterns relevant to
the current anonymous trajectory. AdattTUL \cite{gao2020adversarial} incorporates adversarial transfer
learning into trajectory representation learning to improve the adaptability
of the learned mobility patterns. TGAN \cite{zhou2021improving} employs adversarial trajectory
generation to augment sparse check-in data and enhance the representation of
user mobility patterns. SML-TUL \cite{zhou2021self} applies contrastive learning under spatial
and temporal constraints to learn representations that preserve the consistency
of trajectories generated by the same user.
MainTUL \cite{chen2022mutual} constructs two complementary sequence encoders for the current
trajectory and its augmented historical trajectory. Mobility knowledge is
transferred between the two encoders through mutual distillation, enabling the
current trajectory branch to benefit from long-term historical patterns.
ScaleTUL \cite{zhang2025scalable} further considers large-scale TUL scenarios and develops a
dual-stream representation network. Its short-term and long-term encoders
capture mobility dependencies from trajectories with different temporal
ranges, while spatio-temporal augmentation and representation alignment are
used to improve trajectory representation learning.

These methods demonstrate the effectiveness of recurrent architectures,
attention mechanisms, generative learning, knowledge distillation, and
contrastive learning for extracting sequential user mobility patterns.
Nevertheless, their representations are mainly learned from the POI order and
contextual attributes within individual trajectories. Relations among POIs,
semantic entities, and trajectories are not explicitly represented by a
unified relational structure, which limits the use of global mobility
regularities as prior knowledge for trajectory encoding.

\subsection{Graph-based modeling methods}

Graph-based methods extend trajectory modeling from Euclidean sequences to
non-Euclidean structures. GNNTUL \cite{zhou2021trajectory} constructs a spatially aware POI transition
graph by combining consecutive check-in transitions with geographical
proximity. A graph neural network aggregates neighboring POI information to
learn location representations for trajectory owner prediction. TULRN \cite{sang2023tulrn}
extends this framework by incorporating a prior road network and using
R\'enyi-entropy-based weights to guide neighbor sampling, thereby assigning
different importance to POIs during graph aggregation.

Unlike POI-level graph methods, S2TUL \cite{deng2023s2tul} represents trajectories as graph nodes
and explicitly models inter-trajectory relationships. It constructs
repeatability, spatial-proximity, and spatio-temporal-proximity graphs and
integrates different relation types through graph convolution. The learned
trajectory-level representation is combined with a location-level sequential
representation, while a greedy relinking strategy is used to handle temporal
conflicts among predicted trajectories. AttnTUL \cite{chen2024trajectory} jointly constructs local and
global spatial graphs. Its local graph describes transitions between
geographical regions, whereas its global graph captures relationships among
trajectories and users. Hierarchical spatio-temporal attention is then used to
integrate local sequential dependencies and global trajectory context.
TULMGAT \cite{li2025trajectory} constructs multi-scale check-in-oriented graphs to retain trajectory
context and spatial connectivity. It employs graph attention to model
interactions among check-in nodes and combines representations obtained at
different sampling scales. HGTUL \cite{10.1007/978-981-95-7075-1_21} further introduces a trajectory hypergraph
in which POIs are represented as vertices and trajectories are represented as
hyperedges. This formulation enables a hyperedge to connect multiple POIs and
captures high-order associations among trajectories through shared POI
vertices. A hypergraph attention mechanism is used to distinguish the
contributions of different POIs to each trajectory representation.

The above methods show that graph structures can capture spatial connectivity,
POI transitions, inter-trajectory relationships, and high-order POI
co-occurrence patterns. However, existing graph-based TUL methods primarily
focus on topological connectivity. Visit time, POI category, and mobility
transfer information are generally treated as node attributes or auxiliary
features rather than typed relational knowledge. Consequently, the semantic
roles of different relations are not explicitly distinguished during
trajectory representation learning.

Knowledge graphs provide another graph-based paradigm for organizing
heterogeneous mobility information. Although knowledge graph modeling has not
formed an established research direction in TUL, it has been explored in the
closely related POI recommendation task. Meta-SKR \cite{cui2021sequential} constructs a sequential
knowledge graph that preserves POI check-in order and integrates sequential,
spatio-temporal, and social knowledge for next POI recommendation. Its
knowledge-aware embedding module learns entities and relations while
meta-learning is used to adapt the model to sparse user check-in patterns.
STKGRec \cite{chen2022building} constructs a spatial-temporal knowledge graph directly from user
check-in sequences. It defines spatial-temporal transfer relations between
POIs and exploits knowledge graph representations to model users' long-term
and short-term preferences. Unlike methods that only append temporal and
geographical attributes to POI embeddings, STKGRec organizes mobility
transitions as relational facts and incorporates them into the recommendation
process.
UPSTKGRec \cite{sang2025user} builds a user preference knowledge graph containing
spatio-temporal transfer features. It models POI preferences from both
individual-user and global-user perspectives and constructs user preference
graphs from local knowledge triples. The resulting graph representations are
used to strengthen the connection between users and candidate POIs. DecKG \cite{zheng2025deckg}
introduces knowledge graph enhancement into decentralized POI recommendation.
It pretrains a knowledge graph without exposing private user--POI interactions
and assigns relevant subgraphs to individual users, allowing auxiliary
knowledge to complement limited local check-in data.

These studies indicate that knowledge graphs can organize users, POIs,
categories, temporal contexts, and mobility transitions through explicitly
typed relations. However, their learning objectives are designed to rank or
predict candidate POIs rather than identify the trajectory owner of an
anonymous trajectory. Moreover, they do not consider how multi-relational
mobility embeddings should be integrated with trajectory-level prior
structures and independent user-classification branches for TUL.

Different from existing studies, MakeTUL constructs a multi-relational mobility
knowledge graph containing visit-time, POI-category, and transfer-speed
relations to jointly constrain multimodal mobility embeddings. The knowledge
graph enhanced POI embeddings are processed by the \emph{Trajectory Prior
Structure Extraction Module} to obtain POI representations containing global
structural knowledge. These representations are subsequently integrated with
time, category, and transfer-relation embeddings in the \emph{Trajectory
Sequence Learning Module}, allowing the trajectory representation to capture
both global POI co-occurrence regularities and ordered mobility patterns.
Finally, the \emph{Dual-Branch Classification Layer} retains the global
trajectory representation and the sequence representation in separate
prediction branches and combines their user prediction distributions through
learnable temperature-calibrated fusion.

\section{Preliminaries}

This section introduces the basic concepts used in this paper and formally
defines the TUL problem.

\noindent\textbf{Definition 1 (Check-in Record).}
A check-in record is denoted by $q=(u,p,c,t)$, where $u$ is the user who
generates the record, $p$ is the visited POI, $c$ is the
semantic category of the POI, and $t$ is the corresponding visit timestamp.
Each POI is identified by a unique identifier and is associated with a
geographical coordinate consisting of longitude and latitude.

\noindent\textbf{Definition 2 (Trajectory).}
Let $\mathcal{U}=\{u_1,u_2,\ldots,u_M\}$ denote the set of candidate users,
where $M$ is the number of users. A trajectory generated by a user
$u\in\mathcal{U}$ during a predefined time interval $\tau$ is a sequence of
check-in records arranged in chronological order, denoted by
$\mathcal{S}=(q_1,q_2,\ldots,q_L)$, where $L$ is the trajectory length.
The interval $\tau$ determines how continuous check-in records are divided
into individual trajectories. A trajectory with a known user label is referred
to as a linked trajectory, whereas a trajectory whose generating user is
unknown is referred to as an anonymous trajectory.

\noindent\textbf{Definition 3 (Knowledge Graph).}
A knowledge graph is a structured representation of entities and their
relations. It consists of a set of triples in the form
$\langle h,r,o\rangle$, where $h$ denotes the head entity, $r$ denotes the
relation, and $o$ denotes the tail entity. Each triple expresses a relational
fact indicating that the head entity is connected to the tail entity through
the specified relation.

\noindent\textbf{Problem Formulation.}
Given a set of anonymous trajectories
$\mathcal{T}=\{\mathcal{S}_1,\mathcal{S}_2,\ldots,\mathcal{S}_N\}$
and a set of candidate users
$\mathcal{U}=\{u_1,u_2,\ldots,u_M\}$ who generated these trajectories,
the TUL task aims to learn a classifier that links each anonymous trajectory
to its corresponding trajectory owner:
$\mathcal{T}\mapsto\mathcal{U}$.

\section{Methodology}
\label{sec:methodology}

Figure~\ref{fig:framework} presents the overall architecture of the proposed
MakeTUL. Given a trajectory, the model takes its POI sequence, time
sequence, and category sequence as input. MakeTUL consists of four main
components. First, \emph{Construction of Multi-Relational Knowledge Graph}
organizes visit time, POI category, and movement speed as relational facts.
Second, \emph{Multimodal Embedding Based on Knowledge Graph} learns the
representations of POIs, time entities, category entities, and transfer
relations in relation-specific spaces. Third, \emph{Trajectory Representation
Modeling} extracts structural prior knowledge from the trajectory collection
and learns a sequence representation from prior-enhanced check-ins. Finally,
\emph{User Linking} retains the global trajectory representation and the
sequence representation in two prediction branches and combines their
classification results.

\begin{center}
\centering
    \includegraphics[
    width=\linewidth,
    trim=0.2cm 0.2cm 0.2cm 0.2cm,
    clip
    ]{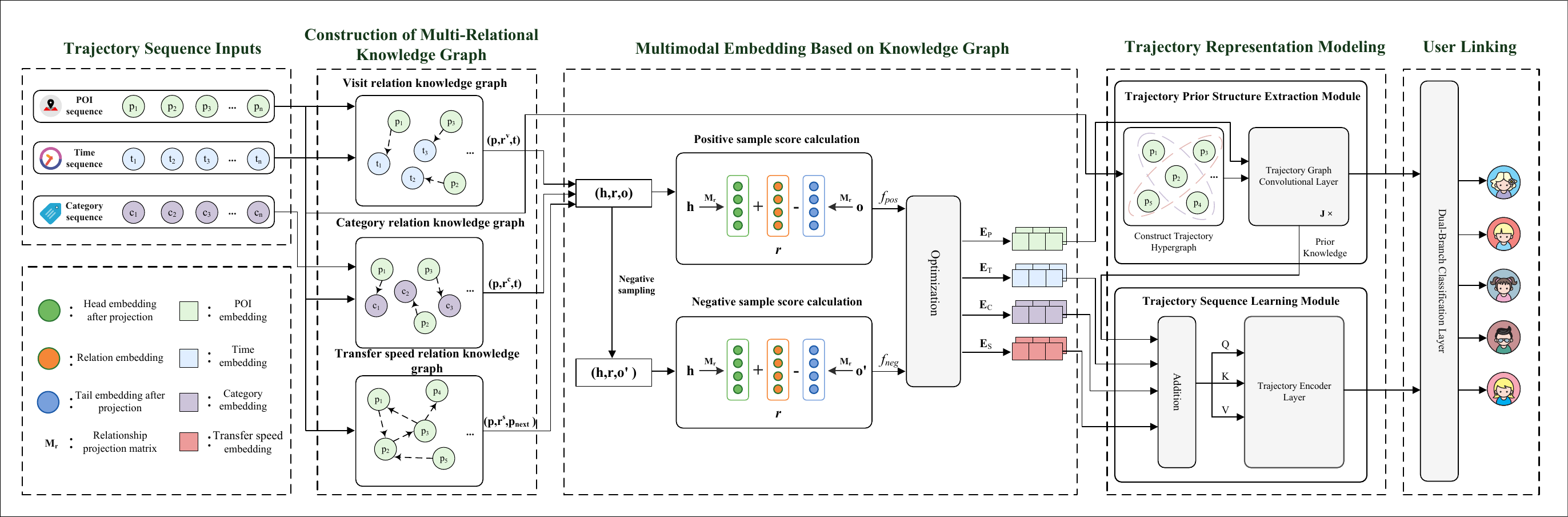}
    \captionof{figure}{Overall architecture of MakeTUL.}
    \label{fig:framework}
\end{center}

\subsection{Construction of multi-relational knowledge graph}
\label{sec:kg_construction}

A check-in trajectory contains several types of mobility information. The POI
sequence records the visited locations, the time sequence describes when these
locations are visited, and the category sequence provides the functional
semantics of the visited POIs. In addition, the movement between consecutive
POIs reflects how a user transfers from one location to another. To organize
these heterogeneous factors in a unified structure, MakeTUL constructs a
multi-relational mobility knowledge graph containing visit-time relations,
POI-category relations, and transfer-speed relations.

\subsubsection{Visit-time relation}

The visit timestamp of each check-in is mapped to one of 48 half-hour time
entities within a day. These time entities retain fine-grained temporal
information. To distinguish the general temporal context of a visit, they are
further divided into four periods. For the $i$-th check-in, the visit-time
relation is defined as

\begin{equation}
r_i^{v}
=
\begin{cases}
r_{\mathrm{midnight}}, & 1\leq \bar{t}_i\leq12,\\
r_{\mathrm{morning}}, & 13\leq \bar{t}_i\leq24,\\
r_{\mathrm{afternoon}}, & 25\leq \bar{t}_i\leq36,\\
r_{\mathrm{evening}}, & 37\leq \bar{t}_i\leq48,
\end{cases}
\label{eq:visit_relation}
\end{equation}

where $\bar{t}_i$ denotes the discretized visit-time entity. Accordingly, the
visit-time information of POI $p_i$ is represented by the triple
$\langle p_i,r_i^{v},\bar{t}_i\rangle$. The time entity preserves the
specific half-hour interval, while the relation indicates the broader period
in which the visit occurs.

\subsubsection{POI-category relation}

Each POI is associated with a semantic category that describes its function,
such as restaurant, office, or shopping venue. The category information of
POI $p_i$ is represented by the triple
$\langle p_i,r^{c},c_i\rangle$, where $r^{c}$ denotes the POI-category
relation and $c_i$ is the category entity associated with $p_i$. This relation
explicitly connects a POI with its functional semantics instead of treating
the POI and its category as unrelated identifiers.

\subsubsection{Transfer-speed relation}

The movement between two consecutive POIs is described by a transfer-speed
relation. For the transition from $p_i$ to $p_{i+1}$, the movement speed is
calculated as

\begin{equation}
v_i
=
\frac{d(p_i,p_{i+1})}
{t_{i+1}-t_i},
\label{eq:transfer_speed}
\end{equation}

where $d(p_i,p_{i+1})$ is the geographical distance between the two POIs,
and $t_{i+1}-t_i$ is the time interval between the corresponding check-ins.
The geographical distance is calculated from the longitude and latitude of
the two POIs using the haversine distance.

The calculated speed is assigned to a predefined speed interval. Let
$\mathcal{B}=\{\beta_0,\beta_1,\ldots,\beta_S\}$ denote the ordered speed
boundaries, where $S$ is the number of speed intervals. For a valid speed
$v_i$, its interval index $b(v_i)$ is determined by

\begin{equation}
b(v_i)=j,\quad
\text{if } \beta_{j-1}\leq v_i<\beta_j,
\qquad j=1,2,\ldots,S.
\end{equation}

The movement from $p_i$ to $p_{i+1}$ is then represented by the triple
$\langle p_i,r_{b(v_i)}^{s},p_{i+1}\rangle$, where
$r_{b(v_i)}^{s}$ denotes the transfer-speed relation associated with the
identified interval. When two consecutive records contain the same POI or
have no valid time difference, the transition is assigned to the first speed
interval.

The three types of triples jointly form the multi-relational mobility
knowledge graph. Visit-time triples describe temporal visiting patterns,
POI-category triples provide functional semantics, and transfer-speed triples
describe movement characteristics between consecutive POIs.

\subsection{Multimodal embedding based on knowledge graph}
\label{sec:kg_embedding}

The constructed knowledge graph contains different types of entities and
relations. A POI can be connected to a time entity through a visit-time
relation, to a category entity through a POI-category relation, and to another
POI through a transfer-speed relation. Since these relations describe
different mobility semantics, representing all entities in a single common
space may weaken their relation-specific meanings. We therefore employ
TransR \cite{lin2015learning} to project entities into the representation space associated with
each relation.

For a positive triple $\langle h,r,o\rangle$, let
$\mathbf{e}_h$, $\mathbf{e}_r$, and $\mathbf{e}_o$ denote the embeddings
of the head entity, relation, and tail entity, respectively. Let
$\mathbf{M}_r$ denote the projection matrix associated with relation $r$.
The positive triple distance is calculated as

\begin{equation}
f_{\mathrm{pos}}(h,r,o)
=
\left\|
\mathbf{M}_r\mathbf{e}_h
+
\mathbf{e}_r
-
\mathbf{M}_r\mathbf{e}_o
\right\|_2^2.
\label{eq:positive_score}
\end{equation}

The projection matrix maps the head and tail entities from the entity space
to the relation-specific space. A smaller value of
$f_{\mathrm{pos}}(h,r,o)$ indicates that the projected head entity combined
with the relation embedding is closer to the projected tail entity.

For each positive triple, $K$ negative triples are constructed by replacing
the original tail entity $o$ with a sampled entity $o'_k$. The distance of
the $k$-th negative triple is

\begin{equation}
f_{\mathrm{neg}}(h,r,o'_k)
=
\left\|
\mathbf{M}_r\mathbf{e}_h
+
\mathbf{e}_r
-
\mathbf{M}_r\mathbf{e}_{o'_k}
\right\|_2^2.
\label{eq:negative_score}
\end{equation}

The negative tail entity is sampled from the entity type allowed by the
current relation. For a visit-time relation, the negative tail is selected
from time entities. For a POI-category relation, it is selected from category
entities. For a transfer-speed relation, it is selected from POI entities.
The sampled entity is required to differ from the original tail entity.

The model is expected to assign a smaller distance to a positive triple than
to its corresponding negative triples. The knowledge graph loss is therefore
defined as

\begin{equation}
\mathcal{L}_{\mathrm{KG}}
=
-\frac{1}{|\mathcal{K}|K}
\sum_{\langle h,r,o\rangle\in\mathcal{K}}
\sum_{k=1}^{K}
\log
\sigma
\left(
f_{\mathrm{neg}}(h,r,o'_k)
-
f_{\mathrm{pos}}(h,r,o)
\right),
\label{eq:kg_loss}
\end{equation}

where $\mathcal{K}$ is the set of positive triples,
$o'_k$ is the negative tail entity of the $k$-th negative triple, and
$\sigma(\cdot)$ is the sigmoid function. Minimizing
Eq.~\eqref{eq:kg_loss} decreases the distances of valid triples and increases
the distances of corrupted triples.

After optimization, the module produces the POI embedding matrix
$\mathbf{E}_P$, the time embedding matrix $\mathbf{E}_T$, the category
embedding matrix $\mathbf{E}_C$, and the transfer-relation embedding matrix
$\mathbf{E}_S$. These representations are jointly constrained by the
multi-relational knowledge graph and are used in the following trajectory
representation modeling process.

\subsection{Trajectory representation modeling}
\label{sec:trajectory_modeling}

Trajectory representation modeling contains two complementary modules:
the \emph{Trajectory Prior Structure Extraction Module} and the
\emph{Trajectory Sequence Learning Module}. The first module extracts
high-order POI co-occurrence information from the trajectory collection. The
second module combines the extracted prior knowledge with temporal, category,
and transfer information to learn a sequence-level trajectory representation.

\subsubsection{Trajectory prior structure extraction module}

A sparse trajectory contains only a limited number of observed check-ins.
Learning its representation from the individual sequence alone may therefore
provide insufficient evidence for distinguishing users with overlapping POI
records. To introduce global structural knowledge, MakeTUL constructs a
trajectory hypergraph in which POIs are treated as nodes and trajectories are
treated as hyperedges. Each hyperedge connects the POIs contained in the
corresponding trajectory.

Let $\mathbf{H}\in\{0,1\}^{N_p\times N_s}$ denote the incidence matrix,
where $N_p$ is the number of POIs and $N_s$ is the number of trajectories.
Its element is defined as

\begin{equation}
H_{ij}
=
\begin{cases}
1, & p_i\in\mathcal{S}_j,\\
0, & \mathrm{otherwise},
\end{cases}
\label{eq:incidence_matrix}
\end{equation}

where $\mathcal{S}_j$ denotes the $j$-th trajectory. Let
$\mathbf{D}_v$ and $\mathbf{D}_e$ denote the node degree matrix and the
hyperedge degree matrix, respectively. Their diagonal elements are

\begin{equation}
(\mathbf{D}_v)_{ii}
=
\sum_{j=1}^{N_s}H_{ij},
\qquad
(\mathbf{D}_e)_{jj}
=
\sum_{i=1}^{N_p}H_{ij}.
\label{eq:hypergraph_degrees}
\end{equation}

The POI embedding matrix learned from the knowledge graph is used as the
initial node representation, i.e.,
$\mathbf{E}_P=\mathbf{P}^{(0)}\in\mathbb{R}^{N_p\times d}$, where $d$ is
the embedding dimension. At the $l$-th trajectory hypergraph convolutional
layer, POI information is first aggregated into trajectory hyperedges and then
propagated back to POI nodes. The propagation is defined as

\begin{equation}
\widetilde{\mathbf{P}}^{(l)}
=
\mathbf{D}_v^{-1}
\mathbf{H}
\mathbf{D}_e^{-1}
\mathbf{H}^{\top}
\mathbf{P}^{(l-1)}
\mathbf{W}^{(l)},
\label{eq:hypergraph_convolution}
\end{equation}

where $\mathbf{W}^{(l)}\in\mathbb{R}^{d\times d}$ is the trainable
transformation matrix of the $l$-th layer. The operation in
Eq.~\eqref{eq:hypergraph_convolution} allows a POI to receive information
from other POIs that occur in the same trajectories.

A residual connection is used to retain the representation from the previous
layer:

\begin{equation}
\mathbf{P}^{(l)}
=
\mathbf{P}^{(l-1)}
+
\widetilde{\mathbf{P}}^{(l)}.
\label{eq:hypergraph_residual}
\end{equation}

Instead of using only the output of the final layer, the representations from
all layers are averaged:

\begin{equation}
\mathbf{P}^{*}
=
\frac{1}{J+1}
\sum_{l=0}^{J}
\mathbf{P}^{(l)},
\label{eq:enhanced_poi}
\end{equation}

where $J$ is the number of trajectory hypergraph convolutional layers.
$\mathbf{P}^{*}$ is the final structure-enhanced POI representation matrix.
It preserves the semantic information learned from the knowledge graph while
incorporating high-order POI co-occurrence information.

The global representations of all trajectories are obtained by aggregating
the enhanced POI representations within their corresponding hyperedges:

\begin{equation}
\mathbf{G}
=
\mathbf{H}^{\top}\mathbf{P}^{*}.
\label{eq:global_trajectory}
\end{equation}

The $j$-th row $\mathbf{g}_j$ of $\mathbf{G}$ denotes the global
representation of trajectory $\mathcal{S}_j$. Meanwhile, the POI-level output
$\mathbf{P}^{*}$ is used as structural prior knowledge in the subsequent
trajectory sequence learning process.

\subsubsection{Trajectory sequence learning module}

For the $i$-th check-in of a trajectory, MakeTUL retrieves the
structure-enhanced POI embedding $\mathbf{p}_i^{*}$ from
$\mathbf{P}^{*}$. The time embedding $\mathbf{t}_i$, category embedding
$\mathbf{c}_i$, and transfer-relation embedding $\mathbf{r}_i^{s}$ are
retrieved from the corresponding rows of $\mathbf{E}_T$, $\mathbf{E}_C$,
and $\mathbf{E}_S$, respectively. Specifically, $\mathbf{t}_i$ represents
the embedding of the visit-time entity associated with $p_i$,
$\mathbf{c}_i$ represents the embedding of its POI category, and
$\mathbf{r}_i^{s}$ represents the embedding of the transfer relation from
$p_i$ to $p_{i+1}$. These representations are added to form the multimodal
check-in representation:

\begin{equation}
\mathbf{x}_i
=
\mathbf{p}_i^{*}
+
\mathbf{t}_i
+
\mathbf{c}_i
+
\mathbf{r}_i^{s}.
\label{eq:checkin_embedding}
\end{equation}

The transfer-relation embedding $\mathbf{r}_i^{s}$ describes the relation
from $p_i$ to $p_{i+1}$. Since the final POI has no subsequent transition,
its transfer-relation embedding is set to a zero vector.

The check-in representations are stacked as

\begin{equation}
\mathbf{X}
=
[\mathbf{x}_1,\mathbf{x}_2,\ldots,\mathbf{x}_L]
\in\mathbb{R}^{L\times d},
\label{eq:trajectory_input}
\end{equation}

where $L$ is the trajectory length and $d$ is the embedding dimension. They
are then processed by stacked trajectory encoder layers based on multi-head
self-attention.

For one attention head, the query, key, and value matrices are calculated as

\begin{equation}
\mathbf{Q}
=
\mathbf{X}\mathbf{W}_Q,
\qquad
\mathbf{K}
=
\mathbf{X}\mathbf{W}_K,
\qquad
\mathbf{V}
=
\mathbf{X}\mathbf{W}_V,
\label{eq:qkv}
\end{equation}

where $\mathbf{W}_Q$, $\mathbf{W}_K$, and $\mathbf{W}_V$ are trainable
projection matrices that map the check-in representations into the query,
key, and value spaces, respectively.

The attention output is calculated as

\begin{equation}
\operatorname{Attn}
(\mathbf{Q},\mathbf{K},\mathbf{V})
=
\operatorname{softmax}
\left(
\frac{\mathbf{Q}\mathbf{K}^{\top}}
{\sqrt{d_h}}
\right)
\mathbf{V},
\label{eq:attention}
\end{equation}

where $d_h$ is the dimension of each attention head. Multiple attention heads
capture different interactions among check-ins. Their outputs are concatenated and projected to obtain the multi-head attention (MHA) output:

\begin{equation}
\operatorname{MHA}(\mathbf{X})
=
\operatorname{Concat}
\left(
\operatorname{head}_1,
\operatorname{head}_2,
\ldots,
\operatorname{head}_A
\right)
\mathbf{W}_O,
\label{eq:multihead_attention}
\end{equation}

where $A$ is the number of attention heads,
$\operatorname{head}_a$ is the output of the $a$-th attention head, and
$\mathbf{W}_O$ is a trainable output projection matrix.

A residual connection and layer normalization are applied to the multi-head
attention output:

\begin{equation}
\mathbf{Z}
=
\operatorname{LayerNorm}
\left(
\mathbf{X}
+
\operatorname{MHA}(\mathbf{X})
\right).
\label{eq:attention_norm}
\end{equation}

The position-wise feed-forward network (FFN) is defined as

\begin{equation}
\operatorname{FFN}(\mathbf{Z})
=
\phi
\left(
\mathbf{Z}\mathbf{W}_1
+
\mathbf{b}_1
\right)
\mathbf{W}_2
+
\mathbf{b}_2,
\label{eq:ffn}
\end{equation}

where $\mathbf{W}_1$ and $\mathbf{W}_2$ are trainable weight matrices,
$\mathbf{b}_1$ and $\mathbf{b}_2$ are bias vectors, and
$\phi(\cdot)$ denotes the ReLU activation function. Another residual
connection and layer normalization are then applied to obtain the encoder
output:

\begin{equation}
\mathbf{Y}
=
\operatorname{LayerNorm}
\left(
\mathbf{Z}
+
\operatorname{FFN}(\mathbf{Z})
\right).
\label{eq:encoder_output}
\end{equation}

Let
$\mathbf{Y}=[\mathbf{y}_1,\mathbf{y}_2,\ldots,\mathbf{y}_L]$
denote the final output of the stacked trajectory encoder layers. A mean
pooling operation is used to obtain the sequence representation:

\begin{equation}
\mathbf{s}
=
\frac{1}{L}
\sum_{i=1}^{L}
\mathbf{y}_i.
\label{eq:mean_pooling}
\end{equation}

The representation is then normalized as

\begin{equation}
\widehat{\mathbf{s}}
=
\frac{\mathbf{s}}
{\|\mathbf{s}\|_2}.
\label{eq:sequence_normalization}
\end{equation}

Through this process, the sequence representation contains the POI transition
information of the current trajectory as well as the structural prior
knowledge extracted from the trajectory collection.

\subsection{User linking}
\label{sec:user_linking}

The global trajectory representation and the sequence representation describe
different aspects of user mobility. The global representation captures
high-order POI co-occurrence patterns across trajectories, whereas the
sequence representation captures the interactions among multimodal check-ins.
MakeTUL retains these two representations as separate prediction branches in
the \emph{Dual-Branch Classification Layer}.

The global representation of the current trajectory is first normalized as

\begin{equation}
\widehat{\mathbf{g}}
=
\frac{\mathbf{g}}
{\|\mathbf{g}\|_2}.
\label{eq:global_normalization}
\end{equation}

The sequence branch and the global branch independently produce user
prediction logits:

\begin{equation}
\mathbf{z}_s
=
\mathbf{W}_s\widehat{\mathbf{s}}
+
\mathbf{b}_s,
\label{eq:sequence_logits}
\end{equation}

\begin{equation}
\mathbf{z}_g
=
\mathbf{W}_g\widehat{\mathbf{g}}
+
\mathbf{b}_g,
\label{eq:global_logits}
\end{equation}

where $\mathbf{W}_s$ and $\mathbf{W}_g$ are trainable classification
matrices, $\mathbf{b}_s$ and $\mathbf{b}_g$ are bias vectors, and $M$ is the
number of candidate users.

The direct combination of the two prediction branches is

\begin{equation}
\mathbf{z}_{\mathrm{plain}}
=
\mathbf{z}_s
+
\mathbf{z}_g.
\label{eq:plain_logits}
\end{equation}

The prediction distributions of the two branches may have different
confidence levels. Two learnable temperature parameters, $T_s$ and $T_g$,
are therefore introduced to calibrate their logits. The calibrated logits are
calculated as

\begin{equation}
\mathbf{z}_{\mathrm{cal}}
=
\frac{\mathbf{z}_s}{T_s}
+
\frac{\mathbf{z}_g}{T_g}.
\label{eq:calibrated_logits}
\end{equation}

A learnable fusion coefficient is defined as

\begin{equation}
\alpha
=
\sigma(a),
\label{eq:fusion_alpha}
\end{equation}

where $a$ is a trainable scalar and
$\sigma(\cdot)$ is the sigmoid function. The sigmoid function
restricts $\alpha$ to the interval $(0,1)$. The final user prediction logits
are obtained by

\begin{equation}
\mathbf{z}
=
(1-\alpha)
\mathbf{z}_{\mathrm{plain}}
+
\alpha
\mathbf{z}_{\mathrm{cal}}.
\label{eq:final_logits}
\end{equation}

The trajectory classification loss is calculated using cross-entropy:

\begin{equation}
\mathcal{L}_{\mathrm{TUL}}
=
-\frac{1}{B}
\sum_{i=1}^{B}
\log
\frac{
\exp(z_{i,y_i})
}{
\sum_{m=1}^{M}
\exp(z_{i,m})
},
\label{eq:tul_loss}
\end{equation}

where $B$ is the batch size, $y_i$ is the ground-truth user label of the
$i$-th trajectory, and $z_{i,m}$ is the prediction logit corresponding to
candidate user $u_m$.

The complete MakeTUL model is trained by jointly optimizing the TUL
classification loss and the knowledge graph loss:

\begin{equation}
\mathcal{L}
=
\mathcal{L}_{\mathrm{TUL}}
+
\lambda_{\mathrm{KG}}
\mathcal{L}_{\mathrm{KG}},
\label{eq:total_loss}
\end{equation}

where $\lambda_{\mathrm{KG}}$ controls the contribution of knowledge graph
representation learning. Joint optimization allows the multi-relational
embeddings, structural prior representations, sequence representations, and
user-linking classifier to be learned for the same TUL objective.

\section{Experiments}
\label{sec:experiments}

In this section, we evaluate MakeTUL through comparative experiments and
model analysis. For comparative experiments, we first introduce the datasets,
evaluation metrics, baselines, and experimental settings, and then compare
MakeTUL with existing TUL methods on three real-world datasets. For model
analysis, we conduct ablation experiments to examine the contribution of the
main components and further investigate the effects of different relations in
the multi-relational knowledge graph. Finally, sensitivity analysis is
performed to study the influence of key model parameters.

\subsection{Datasets and evaluation metrics}
\label{sec:datasets_metrics}

We conduct experiments on three real-world LBSN check-in datasets, namely Foursquare-NYC \cite{yang2014modeling}, Foursquare-TKY \cite{yang2014modeling}, and Foursquare-JKT \cite{yang2015nationtelescope,yang2015participatory}, which contain user check-in records collected in New York City, Tokyo, and Jakarta, respectively. The check-in records are divided into trajectories using one day as the time interval, and users with fewer than
10 trajectories are removed. For each dataset, 20\% of the trajectories are
used as the test set, while the remaining 80\% are used for model development.
We further select 20\% of the training portion as the validation set, with the
remaining trajectories used for training. The validation set is used for
model selection, while the test set is used only for final evaluation.
The detailed statistics of the processed datasets are reported in
Table~\ref{tab:datasets}.

\begin{table}[h]
\centering
\caption{Statistics of the datasets.}
\label{tab:datasets}
\begin{tabular}{ccccc}
\toprule
\textbf{Dataset} & \textbf{Users} & \textbf{Train/Valid/Test} &
\textbf{POIs} & \textbf{Length} \\
\midrule

\multirow{2}{*}{Foursquare-NYC}
& 700
& 43,353/11,188/13,972
& 32,590
& [1, 68] \\

& 1,080
& 59,289/15,371/19,184
& 38,232
& [1, 68] \\

\midrule

\multirow{2}{*}{Foursquare-TKY}
& 700
& 53,687/13,768/17,220
& 37,305
& [1, 78] \\

& 1,400
& 89,821/23,156/28,964
& 51,010
& [1, 78] \\

\midrule

\multirow{2}{*}{Foursquare-JKT}
& 700
& 64,442/16,462/20,580
& 47,041
& [1, 70] \\

& 1,400
& 108,453/27,815/34,771
& 62,680
& [1, 70] \\

\bottomrule
\end{tabular}
\end{table}

Following the commonly used evaluation setting for TUL \cite{zhou2018trajectory,chen2022mutual,deng2023s2tul,li2025trajectory}, we employ five metrics, including ACC@1, ACC@5,
Macro-Precision (Macro-P), Macro-Recall (Macro-R), and Macro-F1. ACC@$k$
measures the proportion of test trajectories whose ground-truth trajectory
owners are ranked within the top-$k$ predicted users, and is calculated as

\begin{equation}
\mathrm{ACC@}k
=
\frac{\#\text{ correctly identified trajectories@}k}
{\#\text{ test trajectories}}.
\end{equation}

In our experiments, both ACC@1 and ACC@5 are reported. Macro-P and Macro-R
are obtained by averaging the precision and recall over all user classes,
respectively, so that each user contributes equally to the evaluation.
Macro-F1 summarizes the balance between Macro-P and Macro-R and is calculated
as

\begin{equation}
\mathrm{Macro\mbox{-}F1}
=
\frac{
2\times\mathrm{Macro\mbox{-}P}\times\mathrm{Macro\mbox{-}R}
}{
\mathrm{Macro\mbox{-}P}+\mathrm{Macro\mbox{-}R}
}.
\end{equation}

\subsection{Baselines}
\label{sec:baselines}

We compare MakeTUL with the following representative TUL baselines:

\begin{itemize}

\item \textbf{TULER \cite{gao2017identifying}}. TULER is an early deep learning framework for the TUL
task. It first learns POI embeddings and then applies RNNS, including Long Short-Term Memory (LSTM), Gated Recurrent Unit (GRU), and bidirectional variants, to model sequential
POI transitions for trajectory classification.

\item \textbf{TULVAE \cite{zhou2018trajectory}}. TULVAE introduces a VAE
into TUL under a semi-supervised learning framework. It uses stochastic latent
variables together with RNN hidden states to capture hierarchical semantics in
check-in trajectories and makes use of unlabeled trajectories to alleviate data
sparsity.

\item \textbf{MainTUL \cite{chen2022mutual}}. MainTUL is a mutual distillation learning framework
designed for sparse check-in trajectories. It employs an RNN encoder to model
the input trajectory and a temporal-aware Transformer encoder to capture
long-term dependencies from augmented historical trajectories. Knowledge is
transferred between the two encoders through mutual distillation.

\item \textbf{S2TUL \cite{deng2023s2tul}}. S2TUL is a semi-supervised graph-based framework that
models both inter-trajectory relations and intra-trajectory sequential
information. It constructs trajectory-level graphs based on repeatability,
spatial proximity, and spatio-temporal proximity, and applies graph neural
networks to learn trajectory-level representations. An LSTM is further used to
model the POI sequence within each trajectory.

\item \textbf{TULMGAT \cite{li2025trajectory}}. TULMGAT constructs check-in-oriented graphs at
multiple sampling scales to preserve the spatio-temporal structure of
trajectories. A masked multi-head graph attention network is used to update
check-in representations, followed by trajectory representation learning for
user linking.

\item \textbf{ScaleTUL \cite{zhang2025scalable}}. ScaleTUL is designed for large-scale TUL and adopts
a spatio-temporal trajectory augmentation strategy together with a dual-stream
representation network. The two streams model long-term and short-term
mobility patterns, while contrastive learning is used to align trajectory and
user representations.

\item \textbf{HGTUL \cite{10.1007/978-981-95-7075-1_21}}. HGTUL introduces hypergraph learning into TUL to model
high-order associations among trajectories. It constructs a trajectory
hypergraph to learn structural trajectory representations and combines them
with spatio-temporal sequence representations for trajectory owner
classification.

\end{itemize}

\subsection{Experimental settings}
\label{sec:experimental_settings}

All experiments are conducted on an NVIDIA GPU, and MakeTUL is implemented
with PyTorch. Adam is adopted for model optimization with a learning rate of
0.001 and a batch size of 128. The embedding dimension for both entities and
relations is set to 128. The trajectory prior structure extraction module uses
2 hypergraph convolutional layers with a dropout rate of 0.3. During knowledge
graph learning, 4 negative samples are generated for each positive triple, and
the weight of the knowledge graph loss $\lambda_{\mathrm{KG}}$ is set to 1.0.
Training lasts for at most 100 epochs, and early stopping is applied with a
patience of 10 epochs. Other parameter settings are summarized in Table~\ref{tab:parameters}.

\begin{table}[h]
\centering
\caption{Parameter settings of MakeTUL.}
\label{tab:parameters}
\begin{tabular}{cc}
\toprule
\textbf{Parameter} & \textbf{Value} \\
\midrule
Model dimension & 128 \\
Learning rate & 0.001 \\
Batch size & 128 \\
Number of hypergraph convolutional layers & 2 \\
Dropout rate & 0.3 \\
Number of negative samples & 4 \\
KG loss weight $\lambda_{\mathrm{KG}}$ & 1.0 \\
\bottomrule
\end{tabular}
\end{table}

\subsection{Experimental results}
\label{sec:experimental_results}

Table~\ref{tab:performance} reports the performance of MakeTUL and the
baseline methods on the three datasets under different user scales. The best
result for each metric is highlighted in bold, while the best baseline result
is underlined. Overall, MakeTUL achieves the best performance on all five
metrics across the six experimental settings. The advantage is observed on
Foursquare-NYC, Foursquare-TKY, and Foursquare-JKT, indicating that the model
maintains stable performance across datasets with different POI distributions
and user scales.

\begin{table}[t]
\centering
\caption{Performance comparison with baselines.}
\label{tab:performance}

\resizebox{\textwidth}{!}{%
\begin{tabular}{cccccccccccc}
\toprule
\multirow{1}{*}{Dataset} & \multirow{1}{*}{Methods}
& ACC@1 & ACC@5 & Macro-P & Macro-R & Macro-F1
& ACC@1 & ACC@5 & Macro-P & Macro-R & Macro-F1 \\
\midrule

% ================= Foursquare-NYC =================
& & \multicolumn{5}{c}{$|U|=700$}
& \multicolumn{5}{c}{$|U|=1080$} \\
\cmidrule(lr){3-7} \cmidrule(lr){8-12}

\multirow{10}{*}{Foursquare-NYC}
& TULER-GRU
& 48.64\% & 55.97\% & 54.84\% & 43.08\% & 44.96\%
& 38.69\% & 45.60\% & 42.51\% & 31.75\% & 32.69\% \\

& TULER-LSTM
& 48.58\% & 55.74\% & 54.00\% & 43.43\% & 45.42\%
& 39.19\% & 45.78\% & 47.40\% & 33.18\% & 36.55\% \\

& BI-TULER
& 51.38\% & 58.81\% & 57.45\% & 46.79\% & 49.26\%
& 42.76\% & 49.01\% & 50.60\% & 37.12\% & 40.14\% \\

& TULVAE
& 49.19\% & 57.56\% & 57.36\% & 45.49\% & 46.09\%
& 42.96\% & 51.35\% & 49.42\% & 38.29\% & 39.04\% \\

& MainTUL
& 56.55\% & 63.13\% & \uline{62.82\%} & 53.89\% & 56.05\%
& 50.69\% & 58.50\% & \uline{56.42\%} & 47.54\% & 49.55\% \\

& S2TUL
& 56.58\% & 63.37\% & 61.57\% & 53.33\% & 54.22\%
& 49.94\% & 57.08\% & 53.27\% & 45.80\% & 45.86\% \\

& TULMGAT
& 36.07\% & 55.39\% & 38.55\% & 33.81\% & 36.03\%
& 29.19\% & 46.88\% & 30.55\% & 26.53\% & 28.40\% \\

& ScaleTUL
& 61.29\% & 68.66\% & 62.03\% & \uline{59.41\%} & \uline{59.22\%}
& \uline{54.64\%} & \uline{62.97\%} & 54.69\%
& \uline{51.74\%} & \uline{51.59\%} \\

& HGTUL
& \uline{62.90\%} & \uline{73.94\%} & 62.48\% & 57.93\% & 57.61\%
& 47.18\% & 62.40\% & 41.41\% & 39.03\% & 36.10\% \\

& Ours
& \textbf{67.83\%} & \textbf{74.51\%} & \textbf{68.48\%}
& \textbf{66.43\%} & \textbf{66.56\%}
& \textbf{60.69\%} & \textbf{68.19\%} & \textbf{60.59\%}
& \textbf{58.45\%} & \textbf{58.39\%} \\

\midrule

% ================= Foursquare-TKY =================
& & \multicolumn{5}{c}{$|U|=700$}
& \multicolumn{5}{c}{$|U|=1400$} \\
\cmidrule(lr){3-7} \cmidrule(lr){8-12}

\multirow{10}{*}{Foursquare-TKY}
& TULER-GRU
& 45.87\% & 59.33\% & 52.66\% & 42.76\% & 44.80\%
& 33.39\% & 44.84\% & 40.16\% & 28.14\% & 29.69\% \\

& TULER-LSTM
& 47.20\% & 60.76\% & 53.87\% & 44.15\% & 46.35\%
& 34.95\% & 47.31\% & 41.37\% & 29.55\% & 31.11\% \\

& BI-TULER
& 49.12\% & 63.26\% & 54.35\% & 45.95\% & 47.45\%
& 36.58\% & 49.56\% & 41.86\% & 31.19\% & 32.40\% \\

& TULVAE
& 50.20\% & 65.33\% & 55.52\% & 48.03\% & 49.22\%
& 39.99\% & 54.31\% & 45.65\% & 35.81\% & 36.08\% \\

& MainTUL
& 59.72\% & 70.15\% & 63.30\% & 57.36\% & 58.26\%
& 51.24\% & 62.17\% & \uline{54.85\%} & 47.51\% & 48.59\% \\

& S2TUL
& 55.88\% & 68.08\% & 60.61\% & 52.44\% & 53.51\%
& 44.07\% & 55.91\% & 45.66\% & 38.67\% & 38.44\% \\

& TULMGAT
& 45.51\% & 67.59\% & 48.49\% & 42.83\% & 45.49\%
& 32.31\% & 53.49\% & 33.69\% & 28.61\% & 30.94\% \\

& ScaleTUL
& \uline{63.99\%} & 74.20\% & \uline{64.19\%}
& \uline{62.19\%} & \uline{61.79\%}
& \uline{53.58\%} & \uline{64.77\%} & 54.13\%
& \uline{50.20\%} & \uline{50.19\%} \\

& HGTUL
& 61.00\% & \uline{77.92\%} & 60.95\% & 54.07\% & 52.76\%
& 35.45\% & 54.69\% & 24.83\% & 25.97\% & 21.03\% \\

& Ours
& \textbf{68.48\%} & \textbf{79.02\%} & \textbf{69.19\%}
& \textbf{66.68\%} & \textbf{67.11\%}
& \textbf{57.72\%} & \textbf{68.88\%} & \textbf{57.72\%}
& \textbf{54.18\%} & \textbf{54.54\%} \\

\midrule

% ================= Foursquare-JKT =================
& & \multicolumn{5}{c}{$|U|=700$}
& \multicolumn{5}{c}{$|U|=1400$} \\
\cmidrule(lr){3-7} \cmidrule(lr){8-12}

\multirow{10}{*}{Foursquare-JKT}
& TULER-GRU
& 42.57\% & 53.44\% & 55.25\% & 39.67\% & 42.59\%
& 32.58\% & 39.86\% & 52.05\% & 28.14\% & 33.05\% \\

& TULER-LSTM
& 44.63\% & 53.61\% & 57.17\% & 41.18\% & 45.37\%
& 33.83\% & 41.34\% & 52.00\% & 29.55\% & 34.32\% \\

& BI-TULER
& 46.43\% & 55.77\% & 57.83\% & 43.04\% & 46.56\%
& 36.19\% & 44.27\% & 52.42\% & 32.01\% & 36.12\% \\

& TULVAE
& 47.06\% & 57.56\% & 57.96\% & 44.80\% & 46.58\%
& 39.58\% & 49.23\% & 49.56\% & 36.48\% & 38.16\% \\

& MainTUL
& 50.46\% & 58.04\% & 59.80\% & 48.53\% & 51.30\%
& 43.49\% & 50.86\% & 55.32\% & 41.41\% & 44.82\% \\

& S2TUL
& 58.39\% & 66.84\% & \uline{65.12\%}
& \uline{55.72\%} & \uline{57.87\%}
& \uline{49.06\%} & \uline{57.73\%} & \uline{55.50\%}
& \uline{45.71\%} & \uline{47.16\%} \\

& TULMGAT
& 31.62\% & 56.19\% & 36.36\% & 29.27\% & 32.43\%
& 21.36\% & 41.80\% & 24.00\% & 18.84\% & 21.11\% \\

& ScaleTUL
& 54.32\% & 62.39\% & 58.55\% & 53.21\% & 54.00\%
& 46.22\% & 54.20\% & 54.31\% & 44.48\% & 46.62\% \\

& HGTUL
& \uline{58.96\%} & \uline{72.60\%} & 60.23\% & 52.57\% & 52.36\%
& 30.32\% & 49.69\% & 22.73\% & 22.12\% & 18.11\% \\

& Ours
& \textbf{67.25\%} & \textbf{74.94\%} & \textbf{70.09\%}
& \textbf{65.66\%} & \textbf{66.65\%}
& \textbf{59.73\%} & \textbf{68.08\%} & \textbf{65.12\%}
& \textbf{58.17\%} & \textbf{59.86\%} \\

\bottomrule
\end{tabular}%
}
\end{table}

Compared with the strongest baseline for each metric, MakeTUL shows consistent
gains across the six settings. On Foursquare-NYC with 700 users, the
improvements in ACC@1, ACC@5, Macro-P, Macro-R, and Macro-F1 are 4.93\%,
0.57\%, 5.66\%, 7.02\%, and 7.34\%, respectively. When the user scale
increases to 1,080, the corresponding gains become 6.05\%, 5.22\%, 4.17\%,
6.71\%, and 6.80\%. Similar results are obtained on Foursquare-TKY. The
largest gains appear on Foursquare-JKT with 1,400 users, where ACC@1 and
Macro-F1 improve by 10.67\% and 12.70\%. These results show that the advantage
of MakeTUL is maintained across different datasets and user scales.

Increasing the number of candidate users makes the TUL task more difficult.
For MakeTUL, ACC@1 decreases from 67.83\% to 60.69\% on Foursquare-NYC,
from 68.48\% to 57.72\% on Foursquare-TKY, and from 67.25\% to 59.73\% on
Foursquare-JKT. Similar decreases can be observed for most baseline methods.
A larger candidate set contains more users with overlapping mobility patterns,
which increases the difficulty of distinguishing trajectory owners. Despite
this change, MakeTUL remains the best-performing method in all larger-scale
settings. The multi-relational knowledge graph and trajectory prior structure
provide additional semantic and structural information, which helps reduce the
ambiguity caused by similar POI visits.

ACC@1 and ACC@5 show different performance patterns. MakeTUL obtains clear
improvements in ACC@1 under all settings, indicating better ranking of the
correct trajectory owner. The improvement in ACC@5 is smaller on
Foursquare-NYC and Foursquare-TKY with 700 users, reaching 0.57\% and 1.10\%,
respectively. HGTUL already obtains high ACC@5 values in these settings, so
the remaining improvement space is limited. MakeTUL therefore maintains
comparable ACC@5 performance while providing a clearer gain in ACC@1. When
the candidate user set becomes larger, its ACC@5 advantage also increases.
For example, the gain reaches 10.35\% on Foursquare-JKT with 1,400 users.

The macro metrics further show that the improvement is not limited to a small
number of users. MakeTUL achieves the highest Macro-P, Macro-R, and Macro-F1
in all settings. On Foursquare-JKT with 1,400 users, Macro-R and Macro-F1
increase by 12.46\% and 12.70\% over the strongest baselines. The
multi-relational knowledge graph provides visit-time, POI-category, and
transfer information, while the trajectory prior structure extraction module
captures POI co-occurrence across trajectories. The trajectory sequence
learning module then combines these representations to model the current
trajectory. Together with the dual-branch classification layer, these
components provide complementary information for trajectory owner
classification, which is consistent with the improvements in both ranking and
macro-averaged metrics.

\subsection{Ablation study}
\label{sec:ablation}

\begin{center}

% ---------------- First row ----------------
\begin{minipage}{0.48\linewidth}
    \centering
    \includegraphics[width=0.96\linewidth]
    {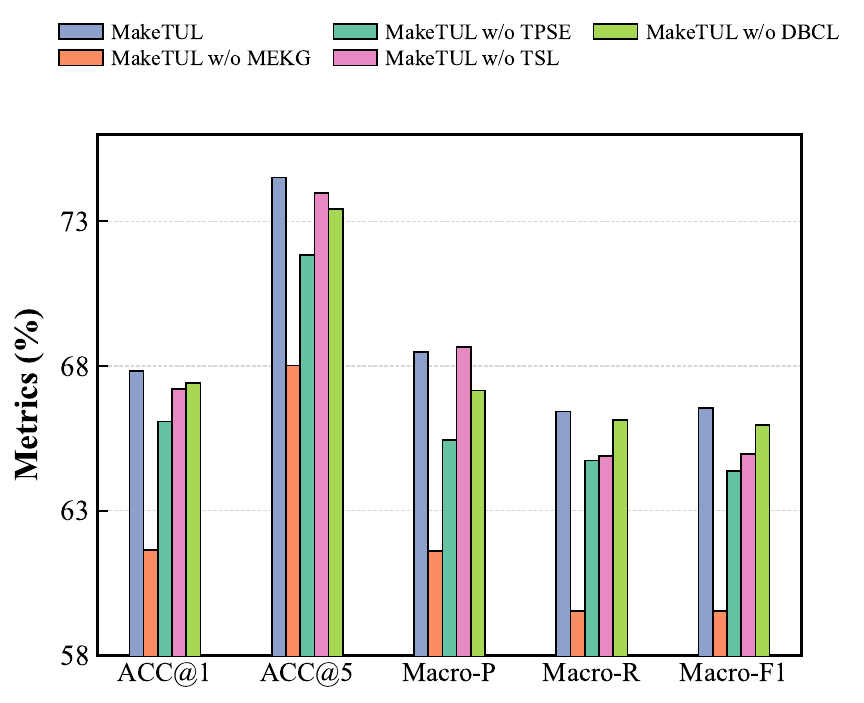}

    \small (a) Foursquare-NYC 700
    \label{fig:NYC700_ablation}
\end{minipage}
\hfill
\begin{minipage}{0.48\linewidth}
    \centering
    \includegraphics[width=0.96\linewidth]
    {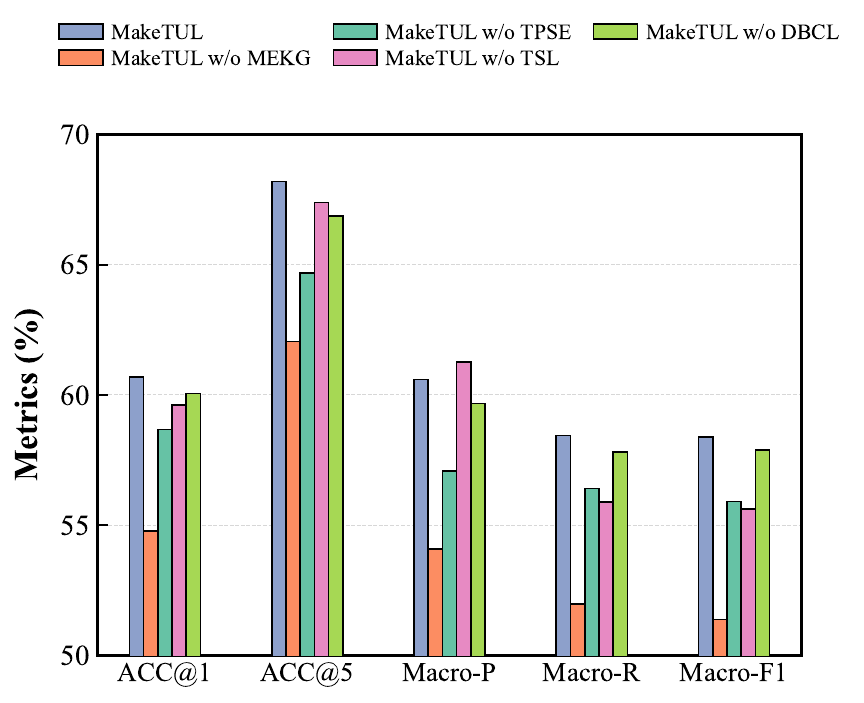}

    \small (b) Foursquare-NYC 1080
    \label{fig:NYC1080_ablation}
\end{minipage}

\medskip

% ---------------- Second row ----------------
\begin{minipage}{0.48\linewidth}
    \centering
    \includegraphics[width=0.96\linewidth]
    {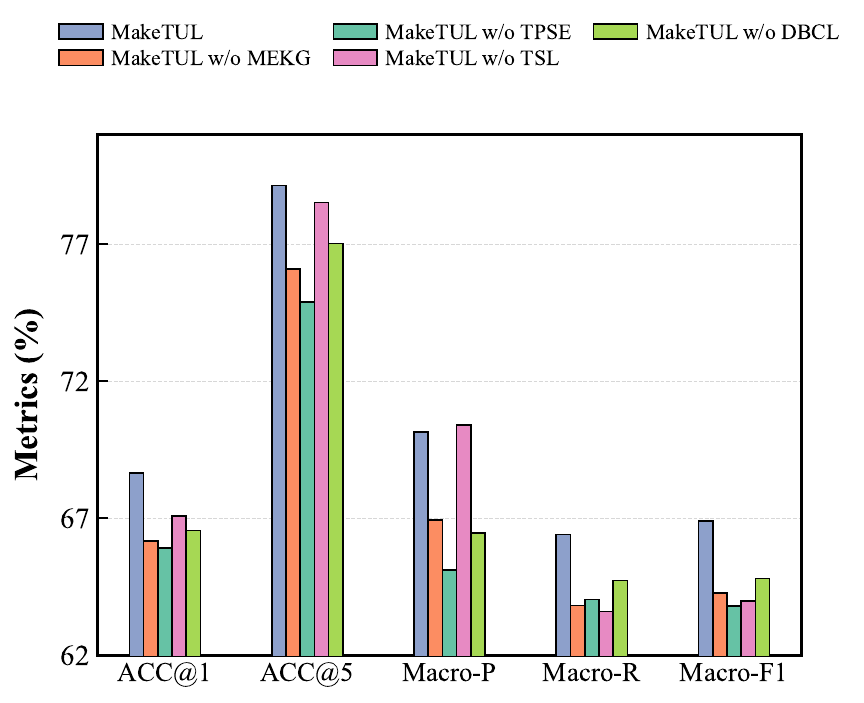}

    \small (c) Foursquare-TKY 700
    \label{fig:TKY700_ablation}
\end{minipage}
\hfill
\begin{minipage}{0.48\linewidth}
    \centering
    \includegraphics[width=0.96\linewidth]
    {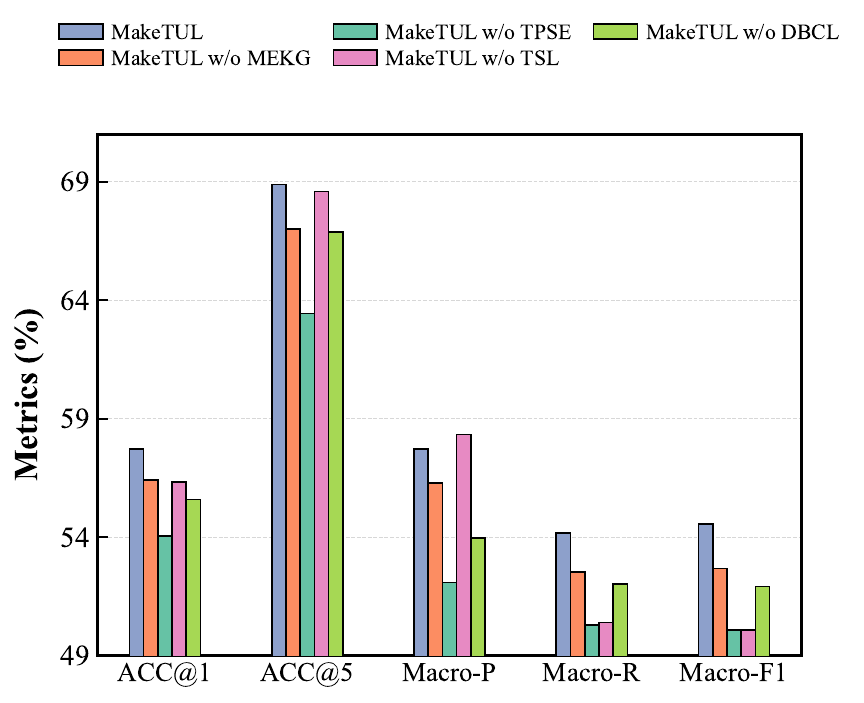}

    \small (d) Foursquare-TKY 1400
    \label{fig:TKY1400_ablation}
\end{minipage}

\medskip

% ---------------- Third row ----------------
\begin{minipage}{0.48\linewidth}
    \centering
    \includegraphics[width=0.96\linewidth]
    {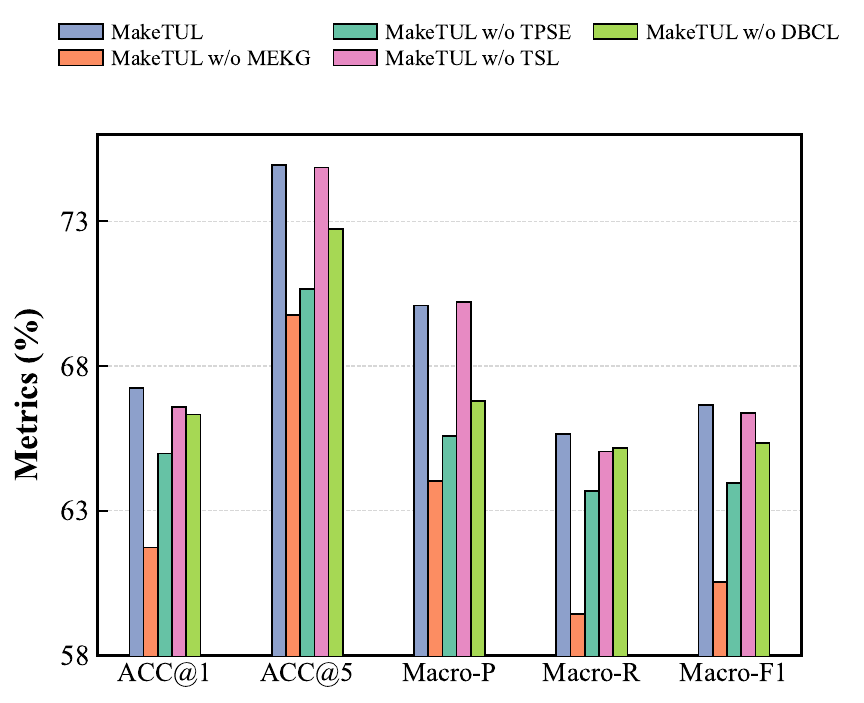}

    \small (e) Foursquare-JKT 700
    \label{fig:JKT700_ablation}
\end{minipage}
\hfill
\begin{minipage}{0.48\linewidth}
    \centering
    \includegraphics[width=0.96\linewidth]
    {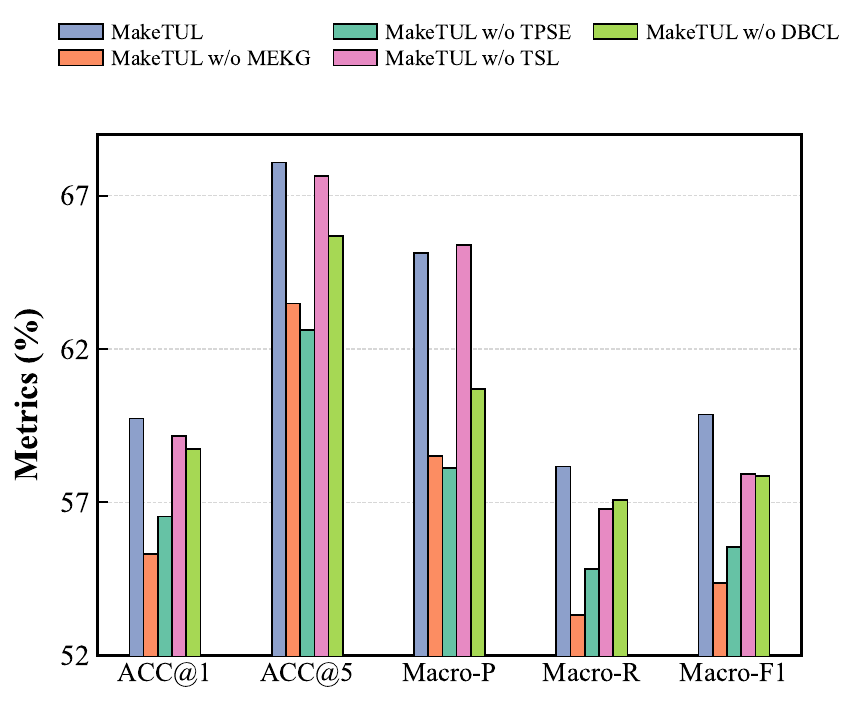}

    \small (f) Foursquare-JKT 1400
    \label{fig:JKT1400_ablation}
\end{minipage}

\captionof{figure}{Ablation results of MakeTUL on different datasets and user scales.}
\label{fig:ablation}

\end{center}

To examine the contribution of each component in MakeTUL, we conduct ablation
experiments on all six experimental settings. Four variants are considered:
MakeTUL w/o MEKG removes the Multimodal Embedding Based on Knowledge Graph
component; MakeTUL w/o TPSE removes the Trajectory Prior Structure Extraction
Module; MakeTUL w/o TSL removes the Trajectory Sequence Learning Module; and
MakeTUL w/o DBCL removes the Dual-Branch Classification Layer. The results are
shown in Fig.~\ref{fig:ablation}.

As shown in Fig.~\ref{fig:ablation}, removing any major component generally
degrades the performance of MakeTUL. The complete model maintains the best
overall results across the six settings, although a few individual metrics of
some variants remain close to or slightly exceed those of the complete model.
This pattern indicates that the four components contribute from different
aspects rather than providing redundant information.

Among the four variants, removing MEKG causes the most evident performance
degradation in most settings. The decreases can be observed not only in ACC@1
and ACC@5 but also in the macro-averaged metrics. Without knowledge graph
embedding, POIs, time information, categories, and transfer relations are no
longer jointly constrained by the multi-relational structure. Consequently,
the trajectory representation loses part of the semantic information that
helps distinguish users who visit similar POIs. The consistent degradation
across NYC, TKY, and JKT supports the role of multi-relational knowledge in
TUL.

Removing TPSE also leads to a clear decrease on most metrics. Without the
trajectory hypergraph, the model mainly relies on information contained in
individual trajectories and cannot fully use POI co-occurrence patterns shared
across trajectories. This effect is particularly relevant when different users
have overlapping POI visits or when an anonymous trajectory contains limited
check-ins. The results therefore show that structural prior knowledge provides
useful complementary information to sequence modeling.

The effect of TSL is relatively moderate compared with MEKG and TPSE, but
removing it still lowers ACC@1, Macro-R, and Macro-F1 in most settings. This
suggests that jointly modeling POI, time, category, and transfer representations
along the trajectory sequence helps refine the trajectory representation. In
several settings, MakeTUL w/o TSL obtains a slightly higher Macro-P than the
complete model. However, this improvement is accompanied by lower recall or
overall classification performance. Therefore, the sequence learning module
mainly contributes to a better balance between precision and recall rather than
optimizing a single metric.

A similar but smaller decline is observed after removing DBCL. The result
shows that the global trajectory representation and the sequence representation
provide complementary evidence for user linking. Using only one source of
trajectory information limits the final classifier, while combining the two
branches produces more stable results across different datasets and user
scales. Overall, the ablation results support the combined use of
multi-relational embedding, trajectory prior structure, sequence learning, and
dual-branch classification in MakeTUL.

\subsection{Analysis of multi-relational knowledge graph}
\label{sec:relation_analysis}

% To examine the contribution of different relations in the multi-relational
% knowledge graph, we construct three variants of MakeTUL. MakeTUL w/o VTR
% removes the visit-time relation, MakeTUL w/o AR removes the attribute relation
% between POIs and their categories, and MakeTUL w/o TR removes the transfer
% relation between consecutive POIs. All other components and experimental
% settings remain unchanged. The results on the six experimental settings are
% shown in Fig.~\ref{fig:relation_analysis}.
To examine the contribution of different relations in the multi-relational
knowledge graph, we construct three variants of MakeTUL. MakeTUL w/o VTR
removes the visit-time relation, MakeTUL w/o PCR removes the
POI-category relation, and MakeTUL w/o TSR removes the
transfer-speed relation. All other components and experimental
settings remain unchanged. The results on the six experimental settings are
shown in Fig.~\ref{fig:relation_analysis}.

As shown in Fig.~\ref{fig:relation_analysis}, the complete MakeTUL generally
achieves more balanced performance across ACC@1 and the macro-averaged
metrics. Removing a single relation changes the results to different degrees,
and the effect varies across datasets and user scales. In most settings,
removing VTR or AR leads to lower overall performance, indicating that both
temporal visiting patterns and POI category semantics provide useful
information for trajectory owner classification. The effect of removing TR is
more dependent on the dataset. Although several metrics decrease after
removing the transfer relation, a few individual results remain comparable to
or slightly higher than those of the complete model.

These results show that the three relations provide complementary information
for TUL. The visit-time relation describes when a POI is visited, which helps
distinguish users with similar visited locations but different temporal
patterns. The attribute relation introduces POI category semantics and
captures differences in users' functional preferences. The transfer relation
describes movement between consecutive POIs and supplements the information
provided by individual visits. Although the contribution of each relation
varies across datasets, combining the three relation types generally produces
more stable performance. This result supports the use of multi-relational
knowledge for modeling user mobility patterns in MakeTUL.

\begin{center}

% ---------------- First row ----------------
\begin{minipage}{0.48\linewidth}
    \centering
    \includegraphics[width=0.85\linewidth]
    {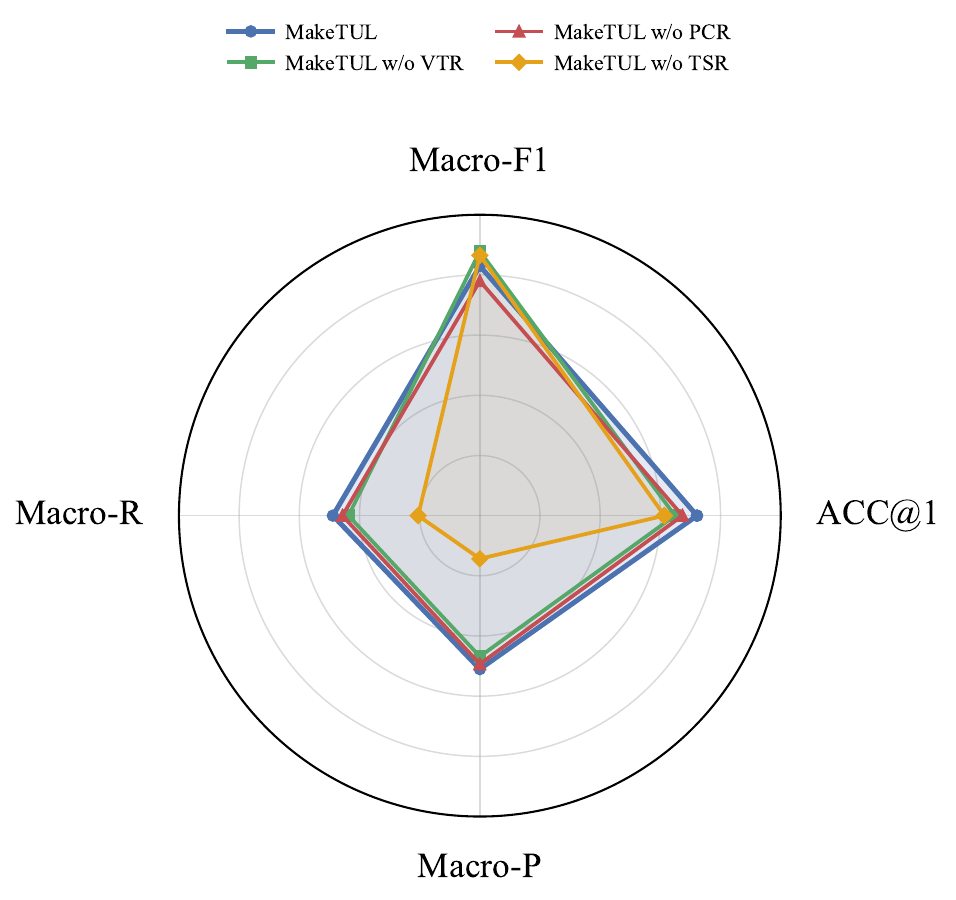}

    \small (a) Foursquare-NYC 700
    \label{fig:relation_NYC700}
\end{minipage}
\hfill
\begin{minipage}{0.48\linewidth}
    \centering
    \includegraphics[width=0.85\linewidth]
    {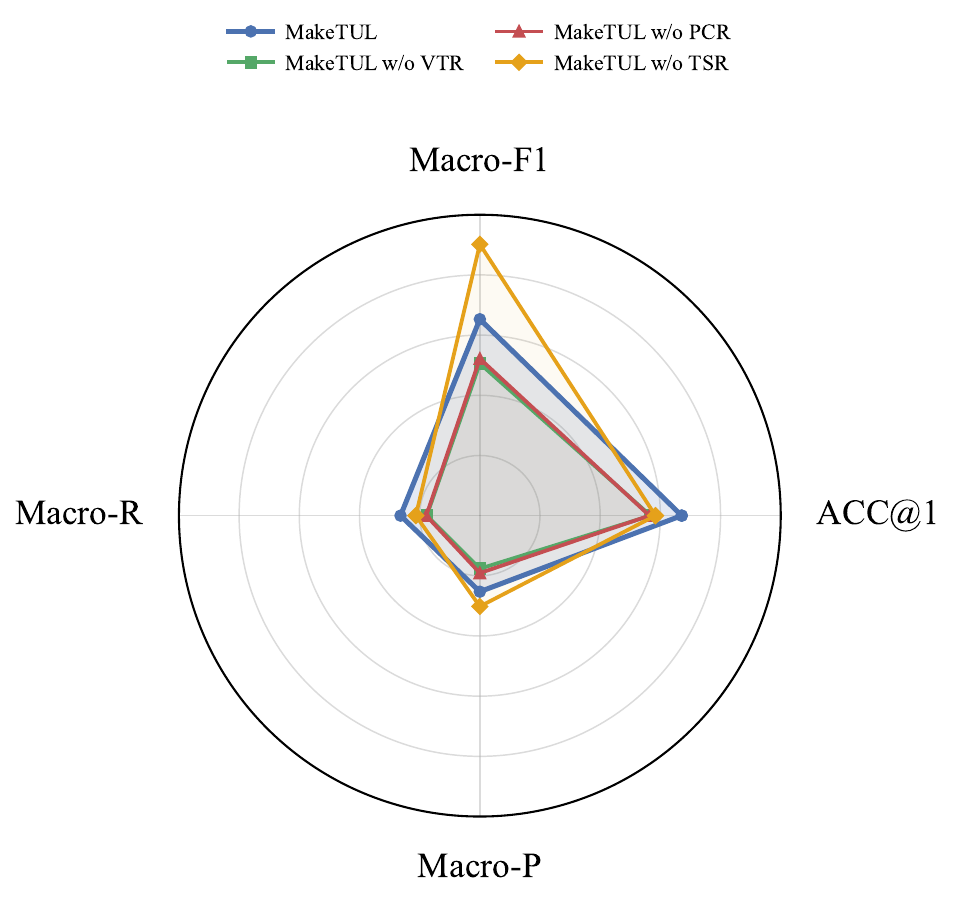}

    \small (b) Foursquare-NYC 1080
    \label{fig:relation_NYC1080}
\end{minipage}

\medskip

% ---------------- Second row ----------------
\begin{minipage}{0.48\linewidth}
    \centering
    \includegraphics[width=0.85\linewidth]
    {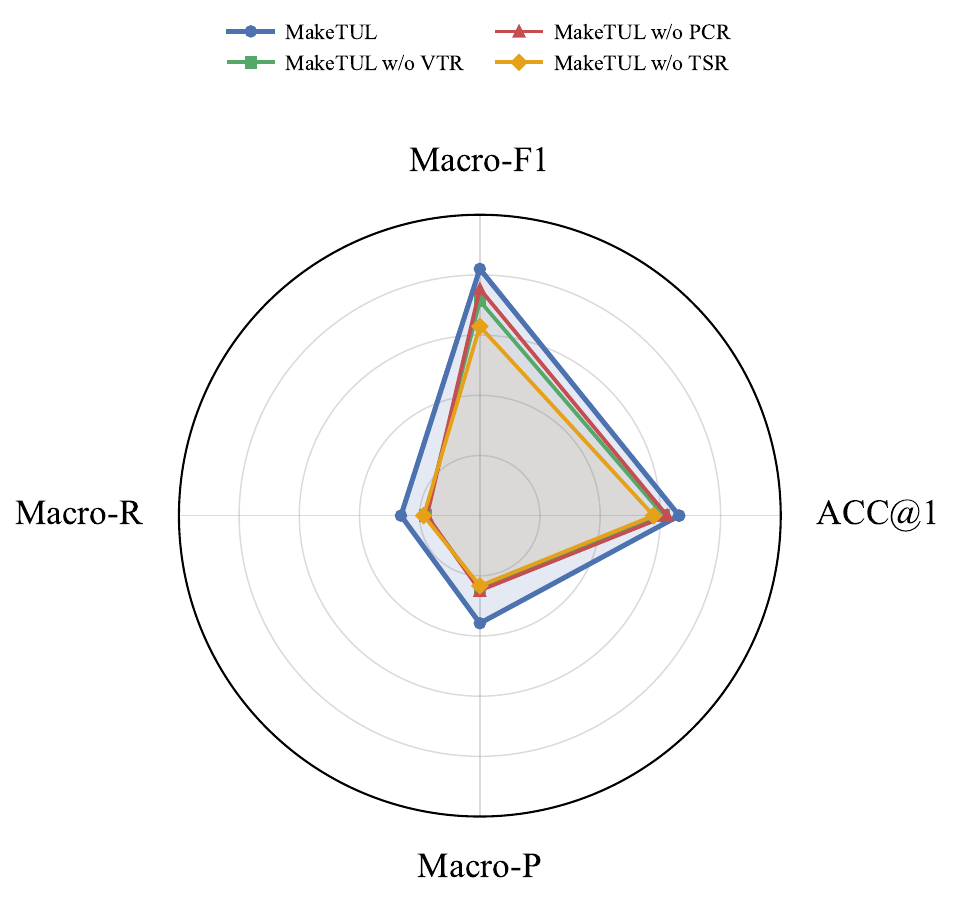}

    \small (c) Foursquare-TKY 700
    \label{fig:relation_TKY700}
\end{minipage}
\hfill
\begin{minipage}{0.48\linewidth}
    \centering
    \includegraphics[width=0.85\linewidth]
    {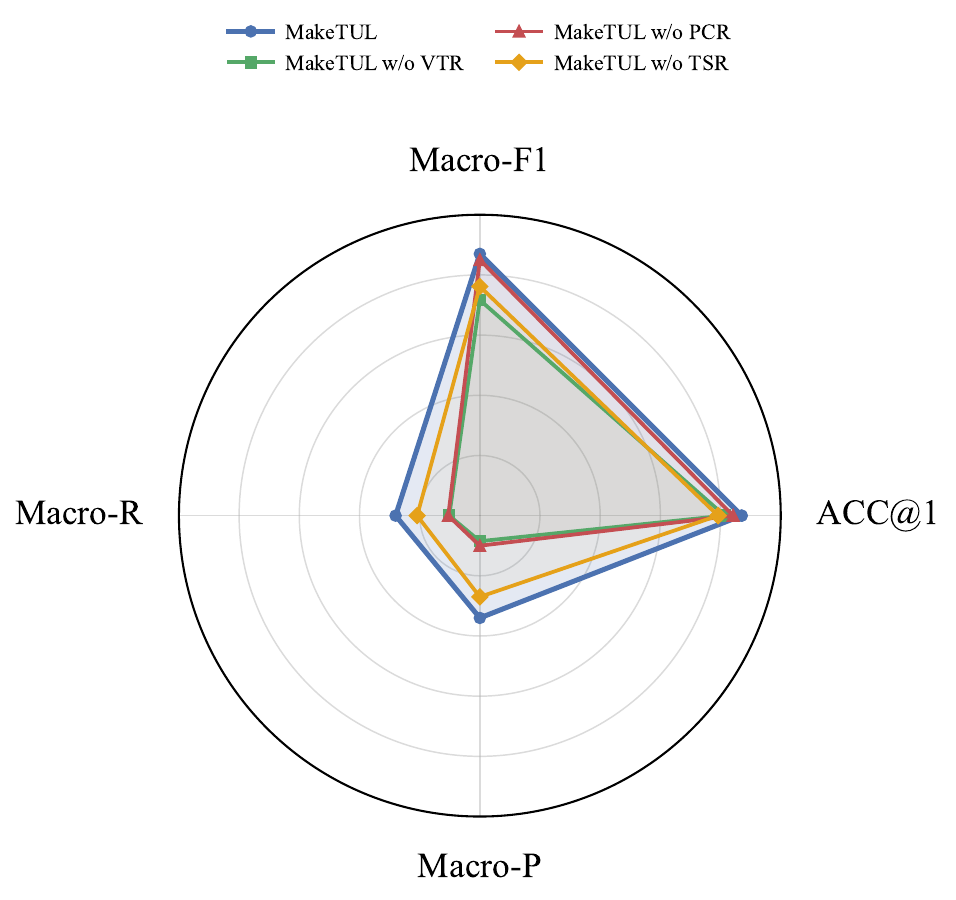}

    \small (d) Foursquare-TKY 1400
    \label{fig:relation_TKY1400}
\end{minipage}

\medskip

% ---------------- Third row ----------------
\begin{minipage}{0.48\linewidth}
    \centering
    \includegraphics[width=0.85\linewidth]
    {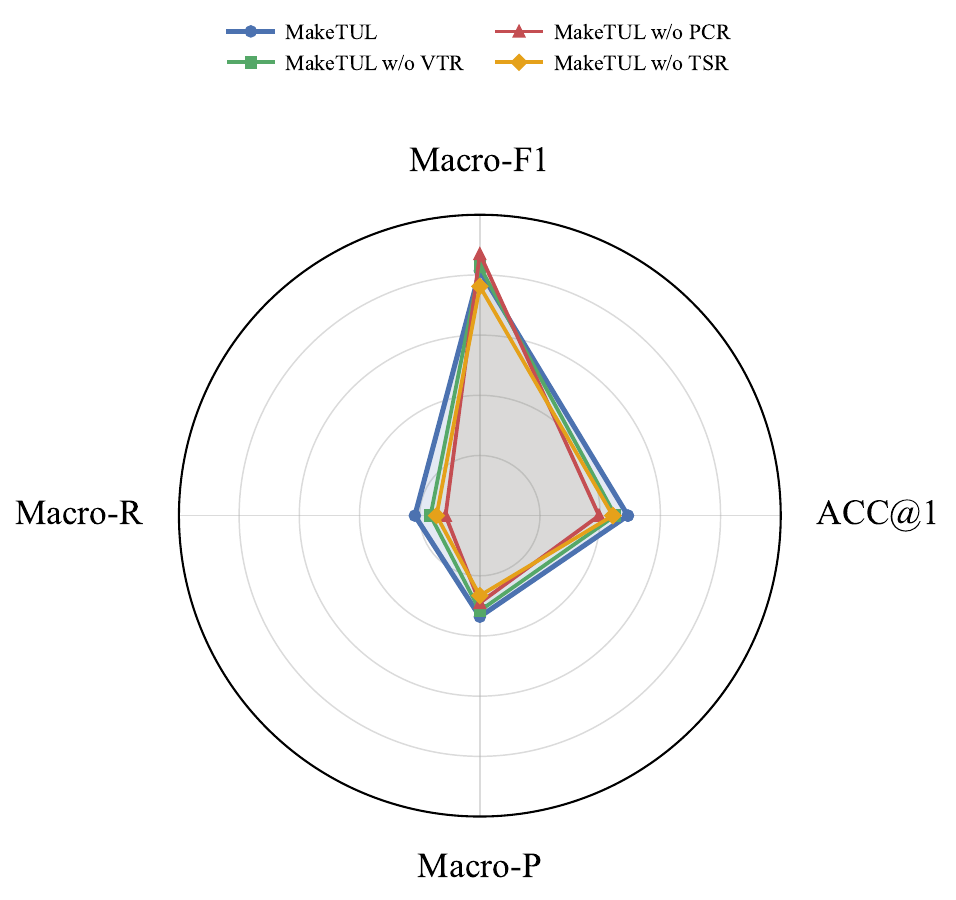}

    \small (e) Foursquare-JKT 700
    \label{fig:relation_JKT700}
\end{minipage}
\hfill
\begin{minipage}{0.48\linewidth}
    \centering
    \includegraphics[width=0.85\linewidth]
    {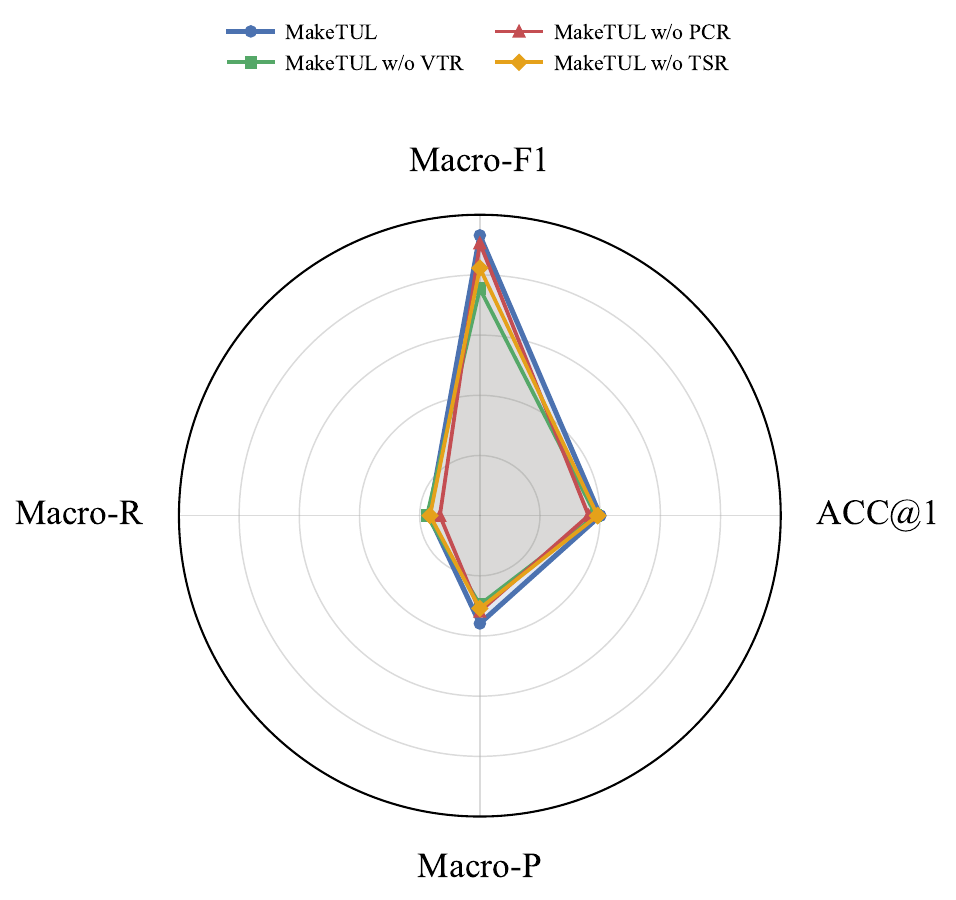}

    \small (f) Foursquare-JKT 1400
    \label{fig:relation_JKT1400}
\end{minipage}

\captionof{figure}{Effects of different relation types in the multi-relational
knowledge graph on different datasets and user scales.}
\label{fig:relation_analysis}

\end{center}

\begin{center}

% ---------------- First row ----------------
\begin{minipage}{0.48\linewidth}
    \centering
    \includegraphics[width=\linewidth]
    {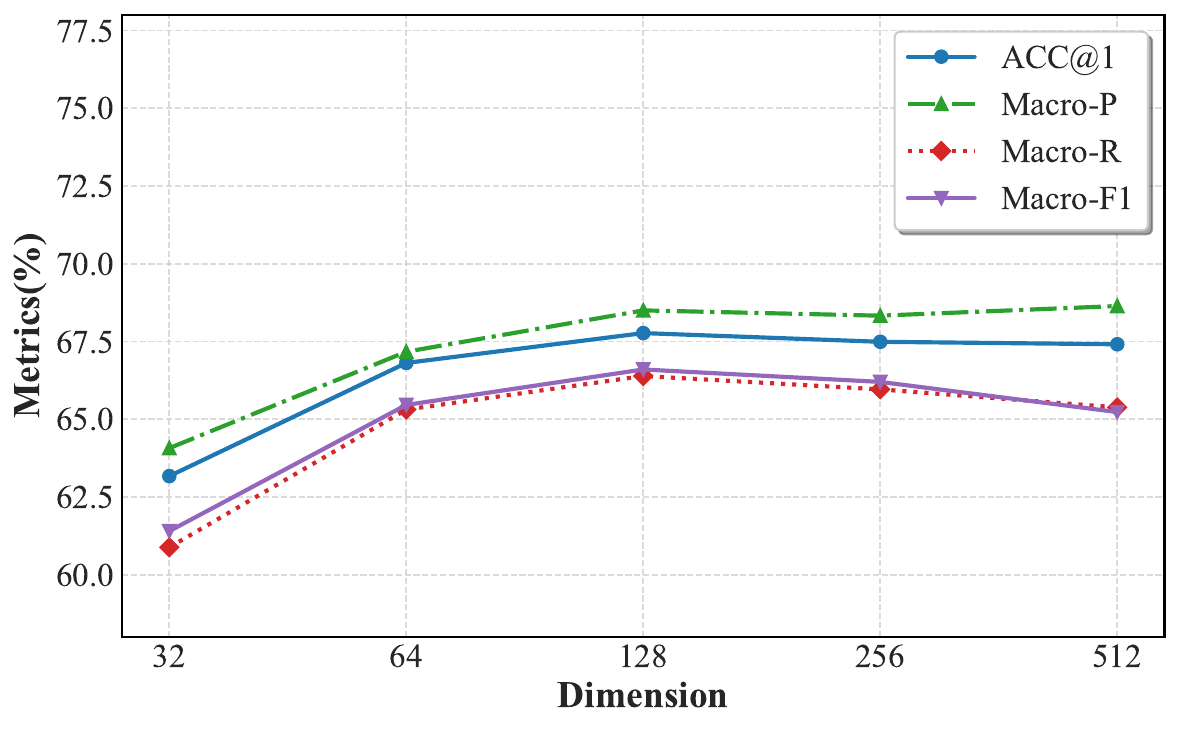}

    \small (a) Foursquare-NYC 700
\end{minipage}
\hfill
\begin{minipage}{0.48\linewidth}
    \centering
    \includegraphics[width=\linewidth]
    {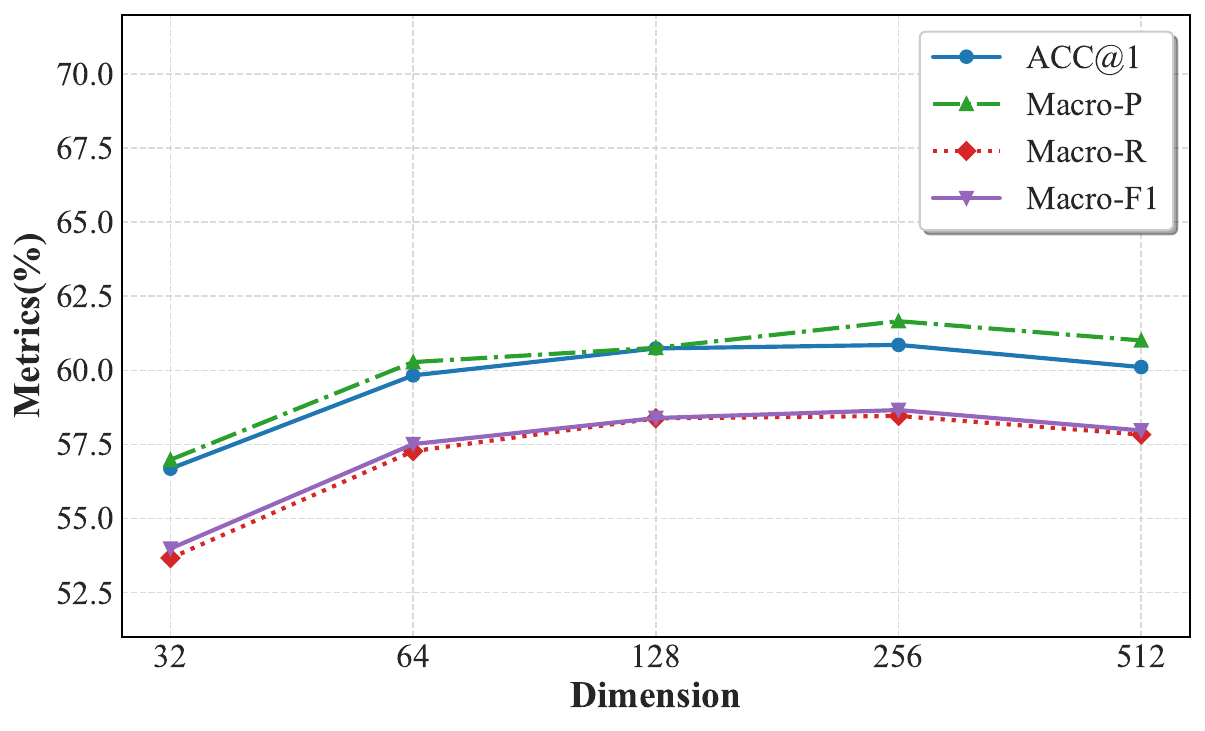}

    \small (b) Foursquare-NYC 1080
\end{minipage}

\medskip

% ---------------- Second row ----------------
\begin{minipage}{0.48\linewidth}
    \centering
    \includegraphics[width=\linewidth]
    {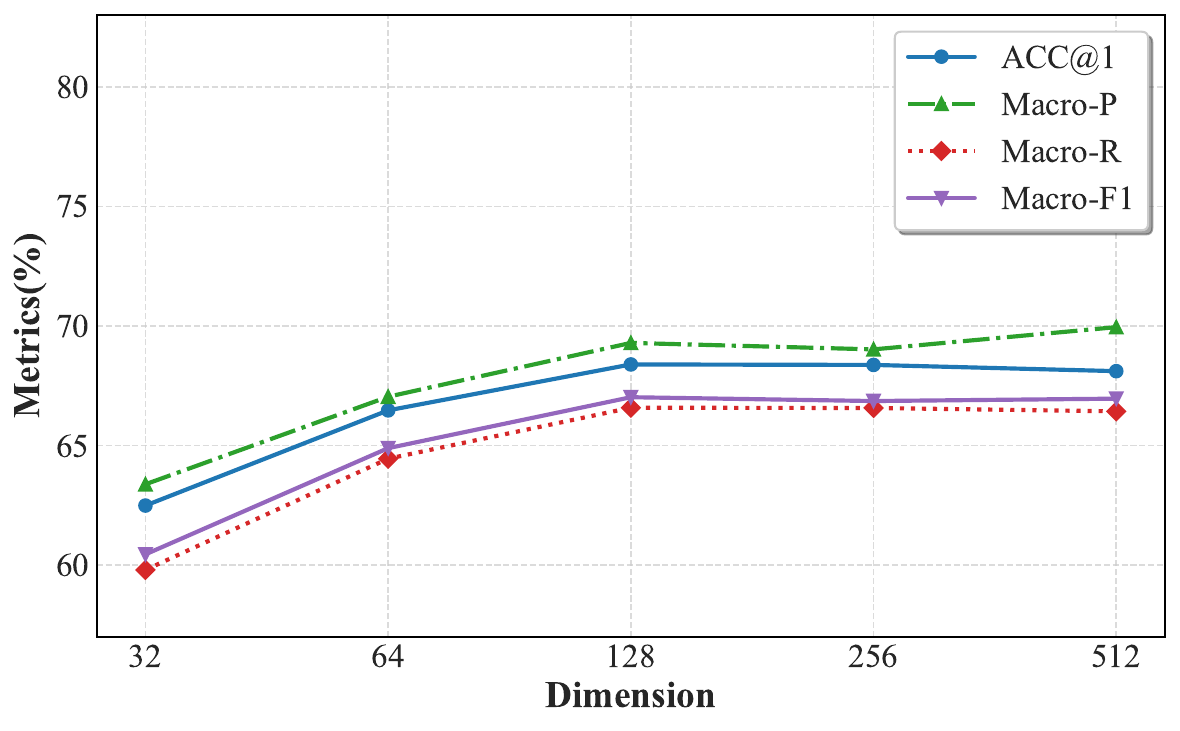}

    \small (c) Foursquare-TKY 700
\end{minipage}
\hfill
\begin{minipage}{0.48\linewidth}
    \centering
    \includegraphics[width=\linewidth]
    {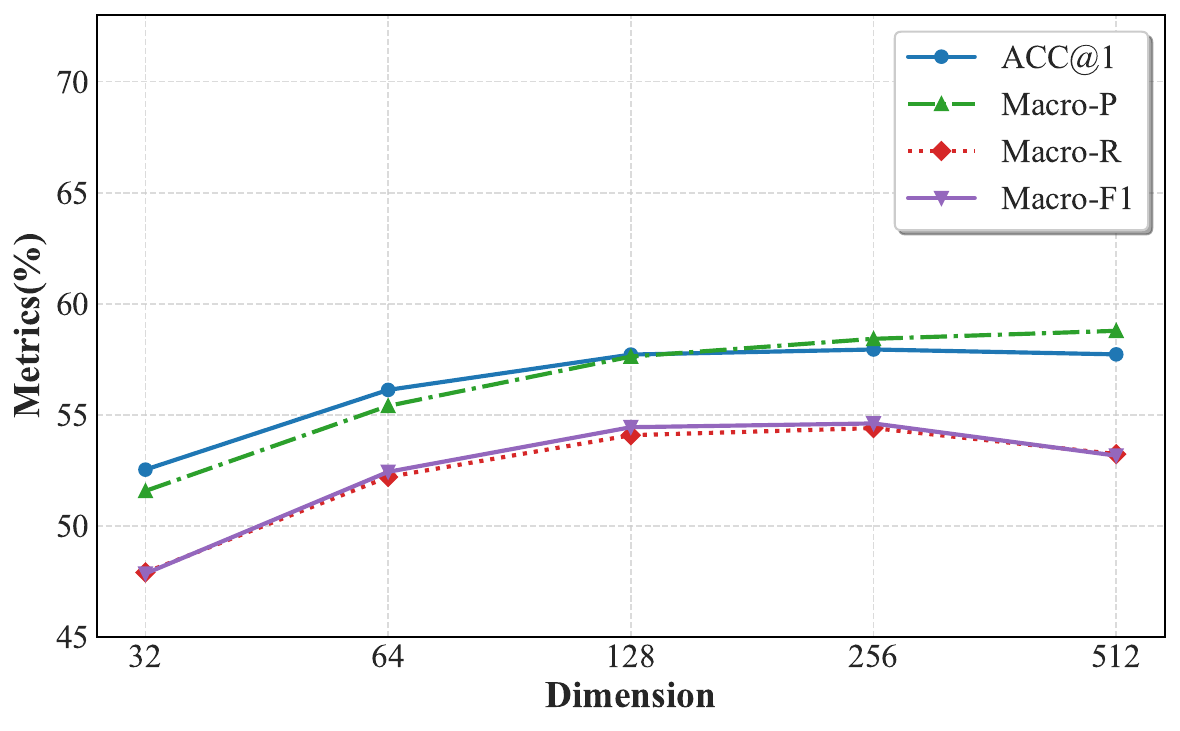}

    \small (d) Foursquare-TKY 1400
\end{minipage}

\medskip

% ---------------- Third row ----------------
\begin{minipage}{0.48\linewidth}
    \centering
    \includegraphics[width=\linewidth]
    {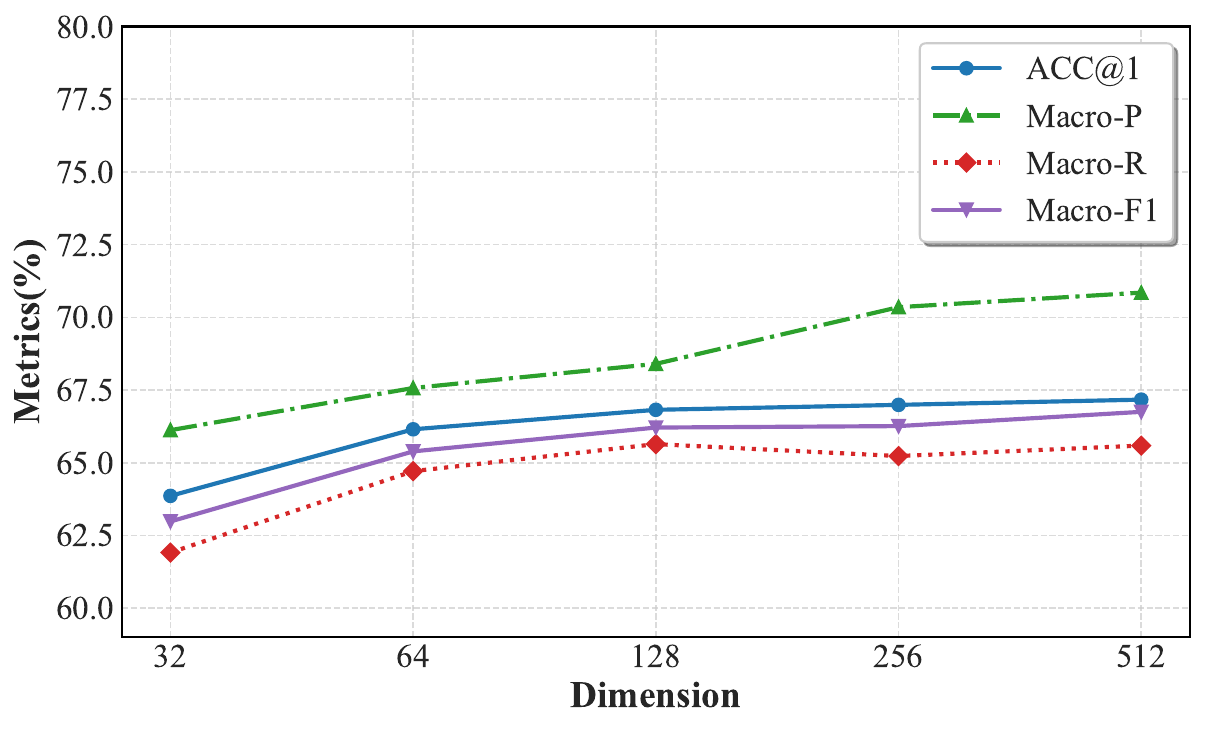}

    \small (e) Foursquare-JKT 700
\end{minipage}
\hfill
\begin{minipage}{0.48\linewidth}
    \centering
    \includegraphics[width=\linewidth]
    {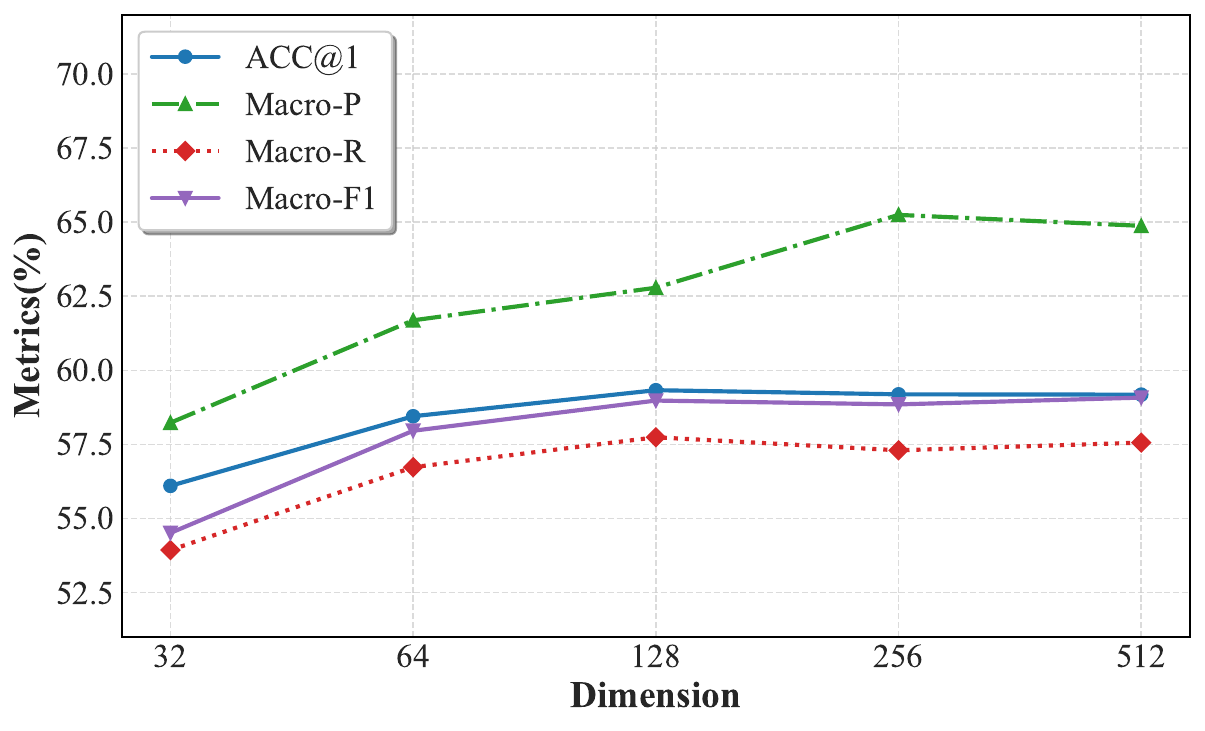}

    \small (f) Foursquare-JKT 1400
\end{minipage}

\captionof{figure}{Sensitivity analysis of the embedding dimension.}
\label{fig:dimension}

\end{center}

\subsection{Sensitivity analysis}
\label{sec:sensitivity}

To examine the sensitivity of MakeTUL to key hyperparameters, we vary the
embedding dimension, the KG loss weight $\lambda_{\mathrm{KG}}$, and the
number of negative samples while keeping the other settings unchanged. The
embedding dimension is selected from $\{32,64,128,256,512\}$. As shown in
Fig.~\ref{fig:dimension}, the performance generally improves as the dimension
increases from 32 to 128. Further increasing the dimension to 256 or 512
provides limited gains and may slightly reduce some metrics. A small dimension
cannot provide sufficient capacity to represent heterogeneous mobility
information, whereas an excessively large dimension introduces additional
model complexity without consistent improvement. Considering both performance
and computational cost, the embedding dimension is set to 128.

\begin{center}

\begin{minipage}{0.48\linewidth}
    \centering
    \includegraphics[width=\linewidth]
    {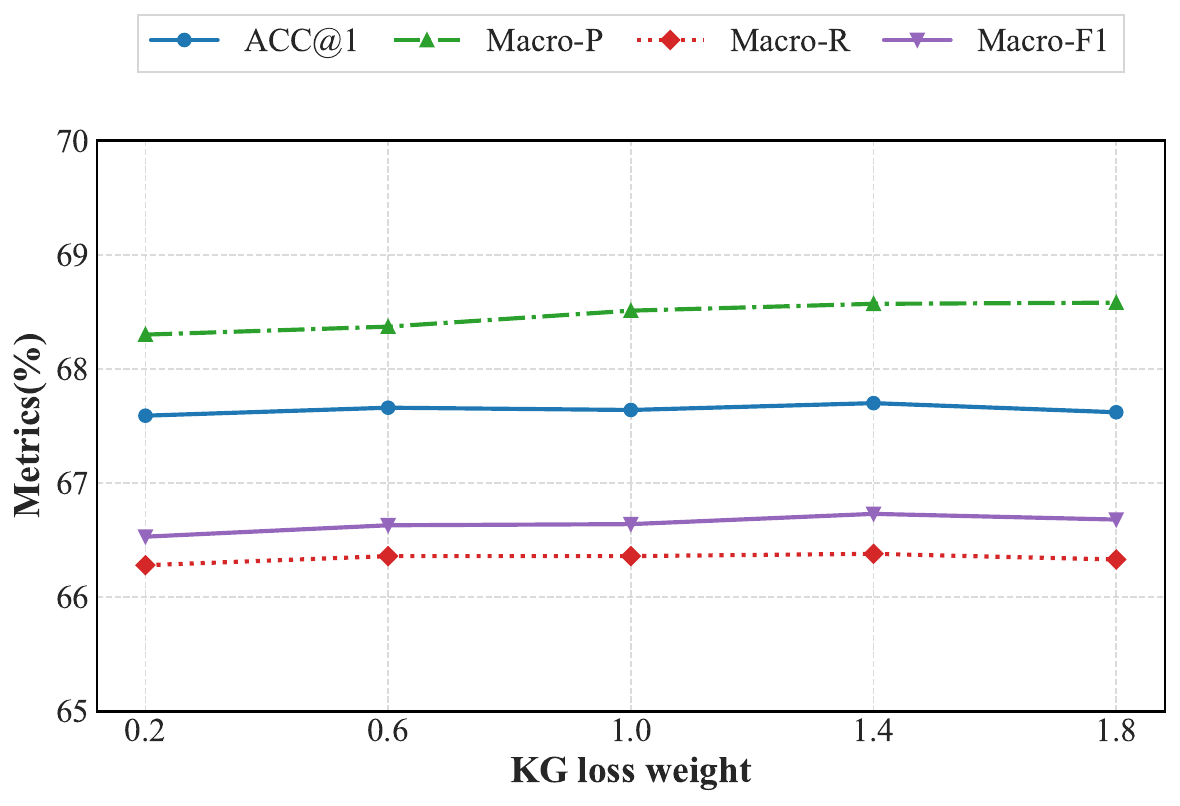}

    \small (a) Foursquare-NYC 700
\end{minipage}
\hfill
\begin{minipage}{0.48\linewidth}
    \centering
    \includegraphics[width=\linewidth]
    {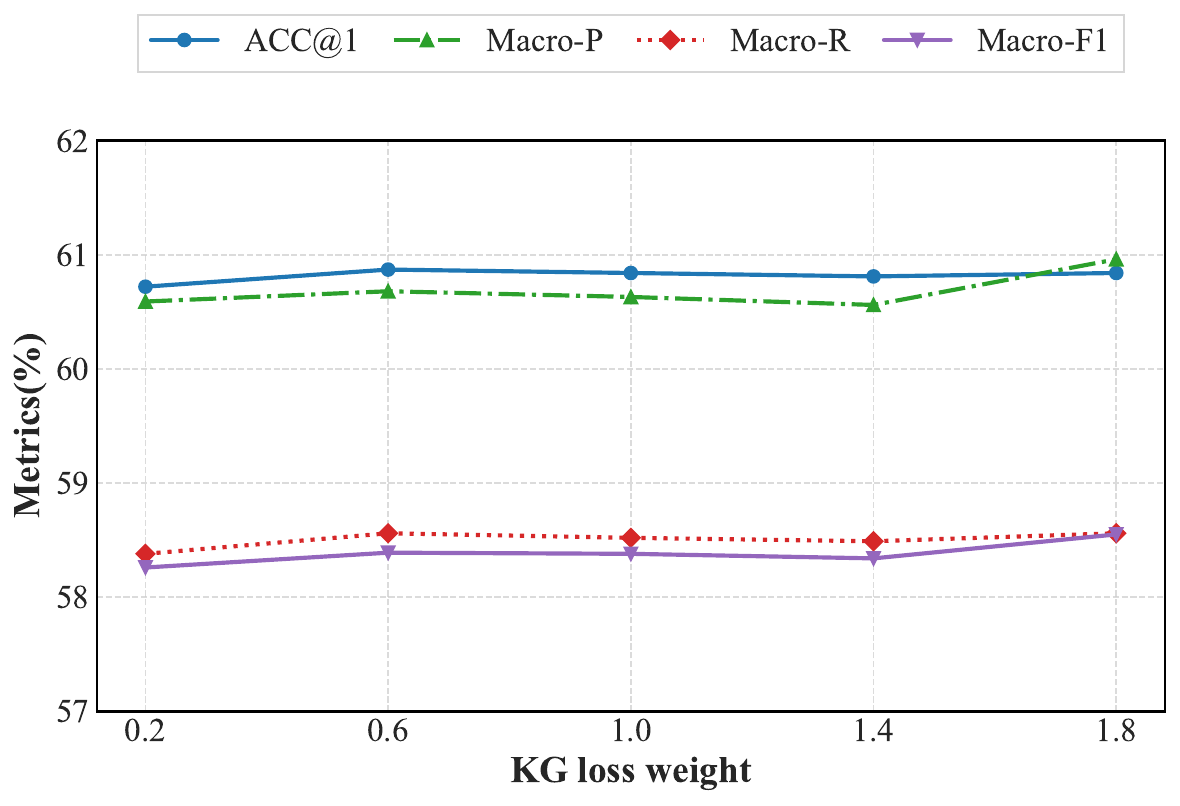}

    \small (b) Foursquare-NYC 1080
\end{minipage}

\medskip

\begin{minipage}{0.48\linewidth}
    \centering
    \includegraphics[width=\linewidth]
    {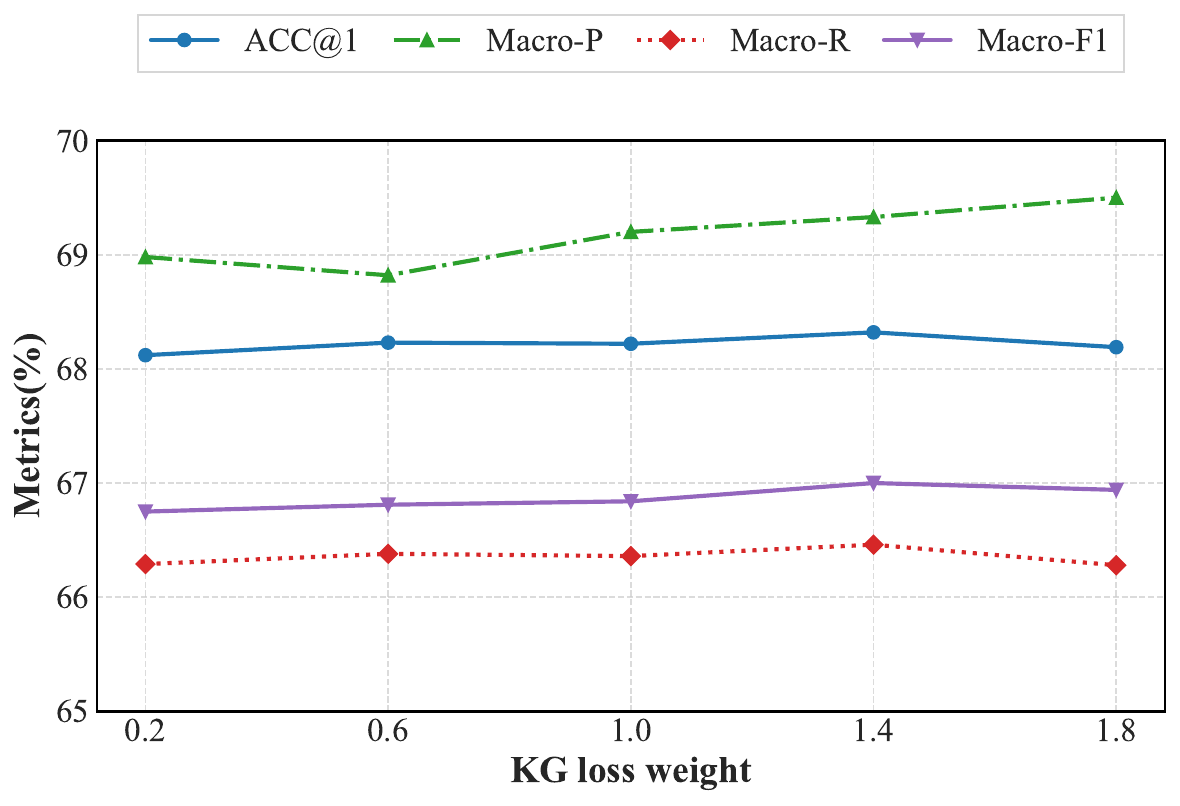}

    \small (c) Foursquare-TKY 700
\end{minipage}
\hfill
\begin{minipage}{0.48\linewidth}
    \centering
    \includegraphics[width=\linewidth]
    {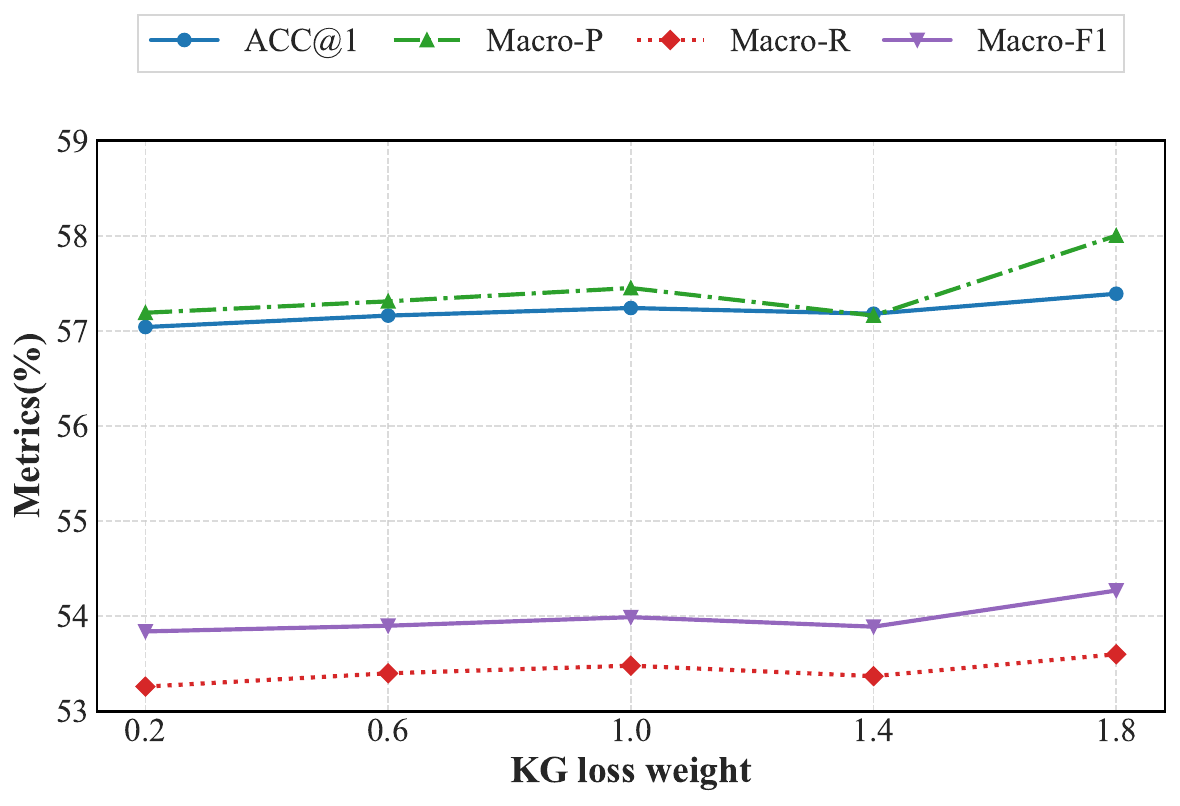}

    \small (d) Foursquare-TKY 1400
\end{minipage}

\medskip

\begin{minipage}{0.48\linewidth}
    \centering
    \includegraphics[width=\linewidth]
    {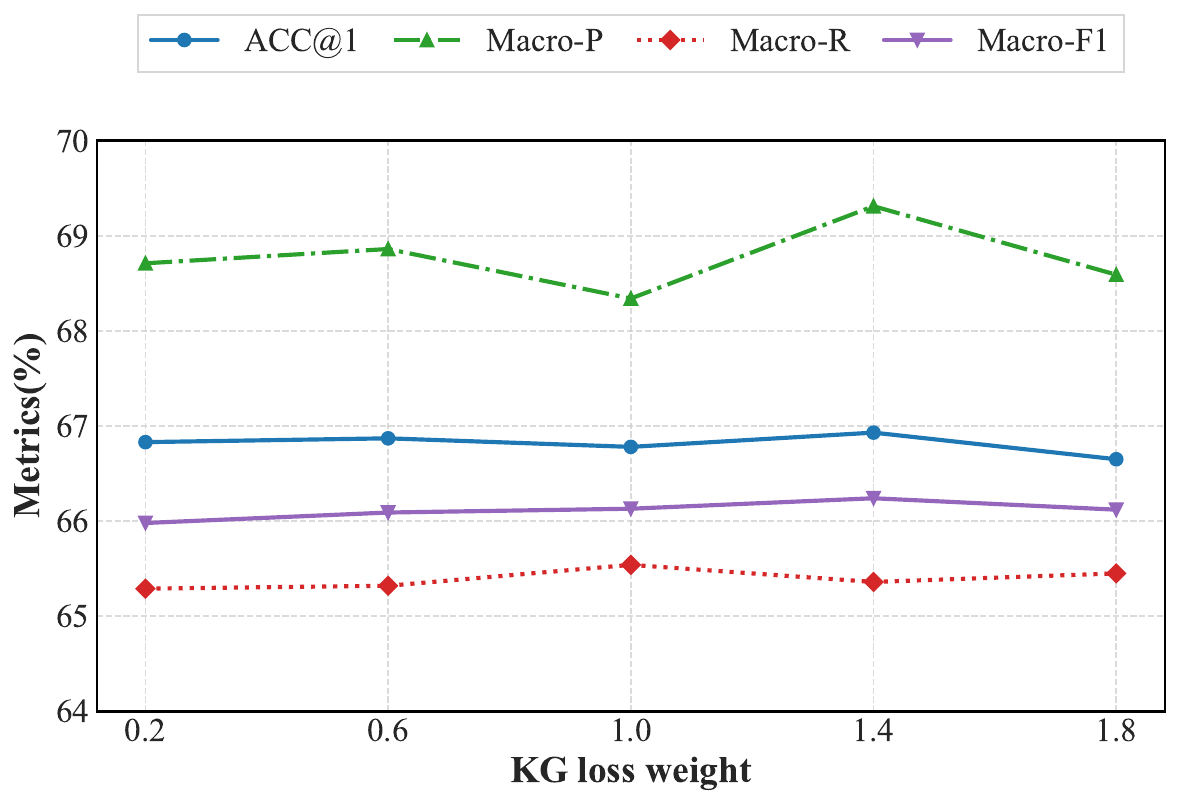}

    \small (e) Foursquare-JKT 700
\end{minipage}
\hfill
\begin{minipage}{0.48\linewidth}
    \centering
    \includegraphics[width=\linewidth]
    {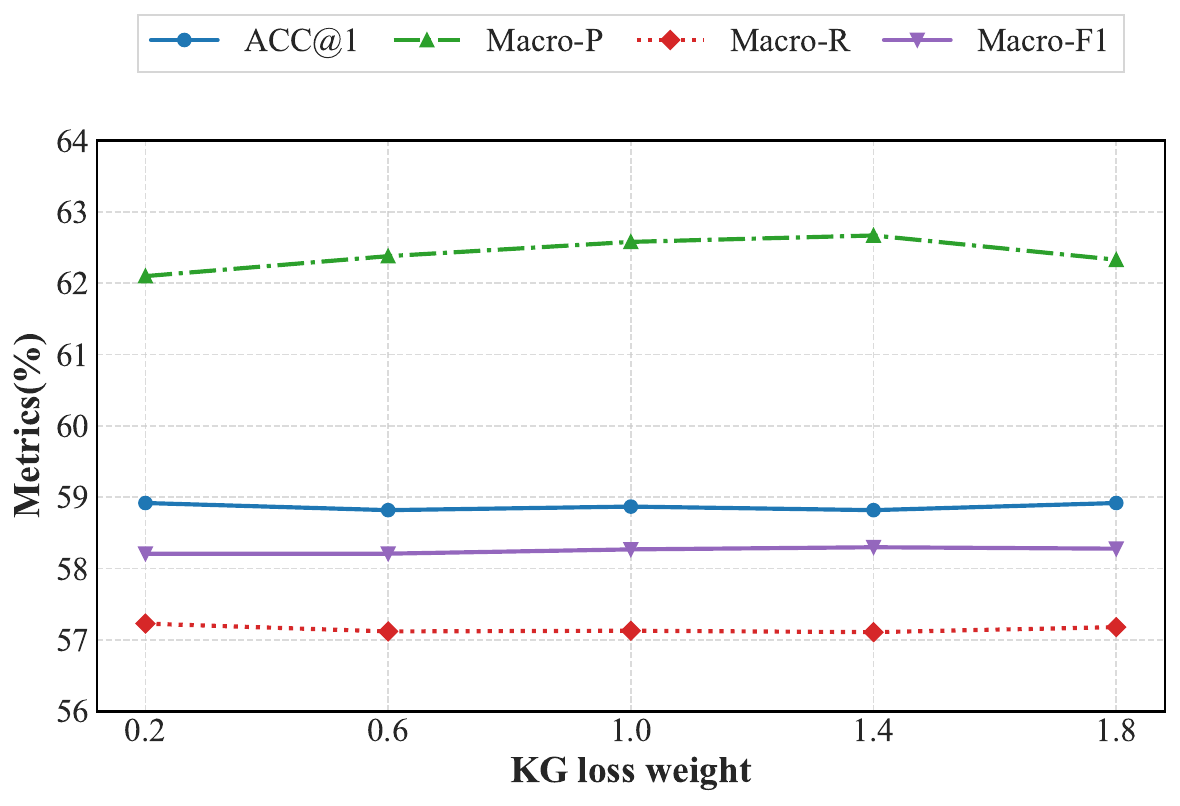}

    \small (f) Foursquare-JKT 1400
\end{minipage}

\captionof{figure}{Sensitivity analysis of the KG loss weight
$\lambda_{\mathrm{KG}}$.}
\label{fig:kg_weight}

\end{center}

The effect of $\lambda_{\mathrm{KG}}$ is presented in
Fig.~\ref{fig:kg_weight}. We vary its value from 0.2 to 1.8. The performance
remains relatively stable across this range on all datasets, although the best
value of individual metrics varies slightly among different settings. This
result indicates that MakeTUL is not highly sensitive to the weight of the
knowledge graph objective. A value of 1.0 provides competitive and stable
performance while maintaining a balanced contribution between trajectory
classification and knowledge graph representation learning.

\begin{center}

\begin{minipage}{0.48\linewidth}
    \centering
    \includegraphics[width=\linewidth]
    {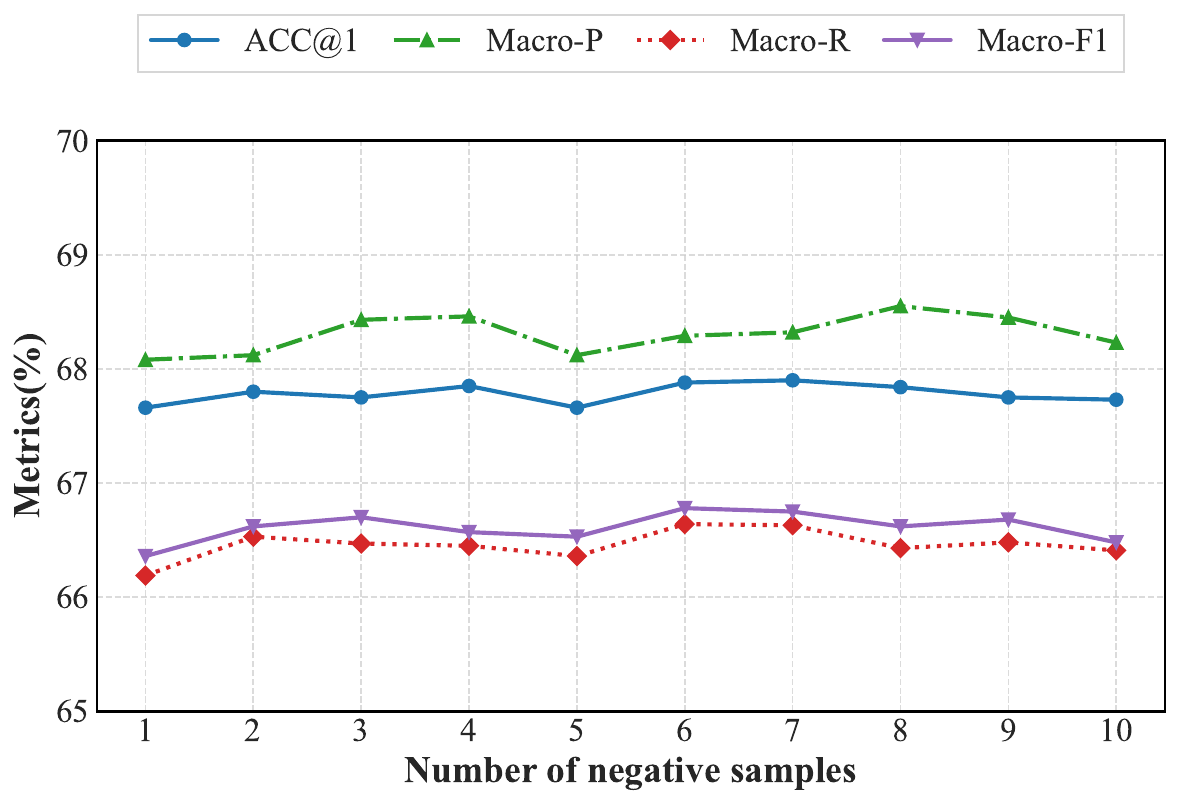}

    \small (a) Foursquare-NYC 700
\end{minipage}
\hfill
\begin{minipage}{0.48\linewidth}
    \centering
    \includegraphics[width=\linewidth]
    {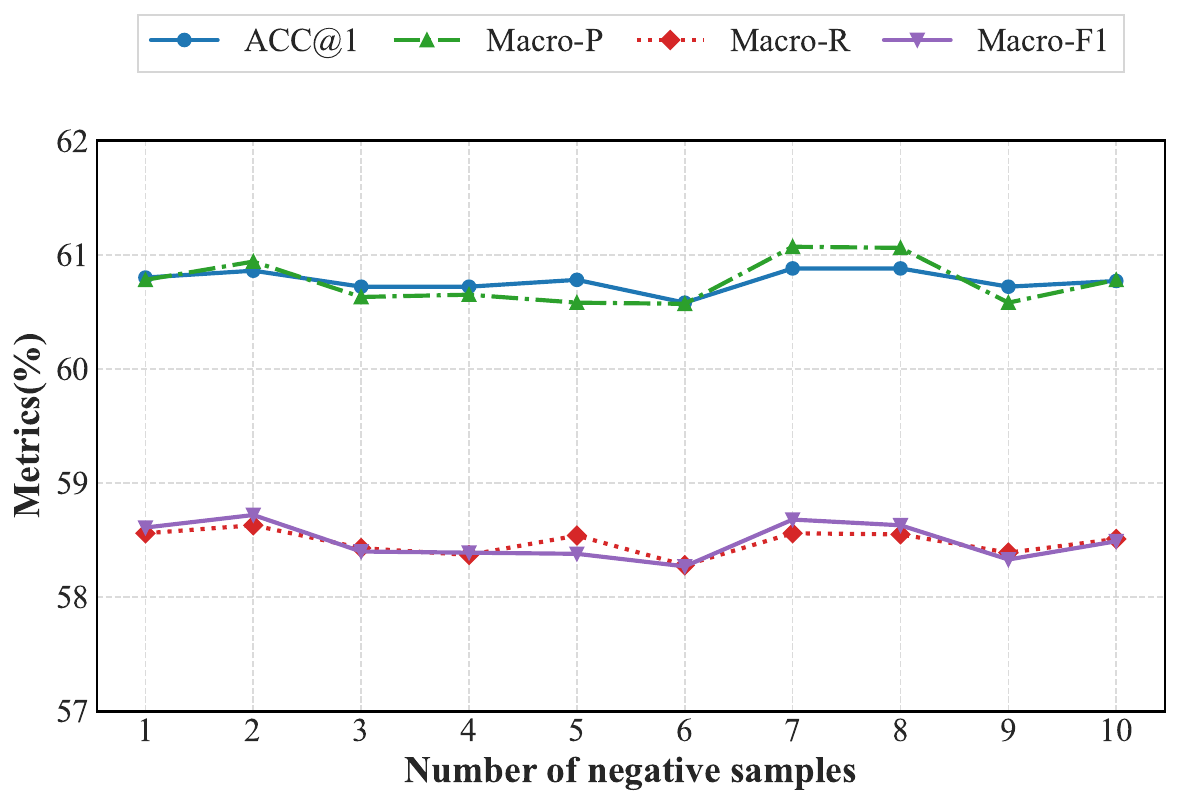}

    \small (b) Foursquare-NYC 1080
\end{minipage}

\medskip

\begin{minipage}{0.48\linewidth}
    \centering
    \includegraphics[width=\linewidth]
    {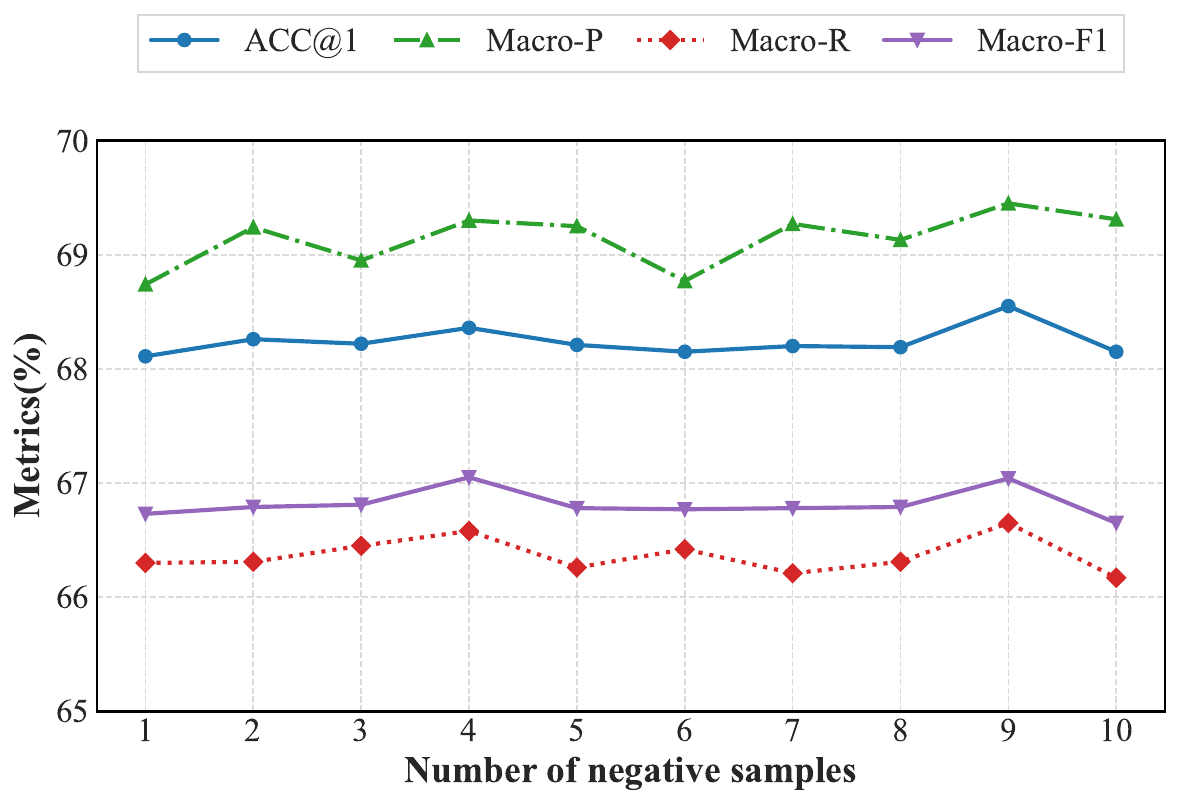}

    \small (c) Foursquare-TKY 700
\end{minipage}
\hfill
\begin{minipage}{0.48\linewidth}
    \centering
    \includegraphics[width=\linewidth]
    {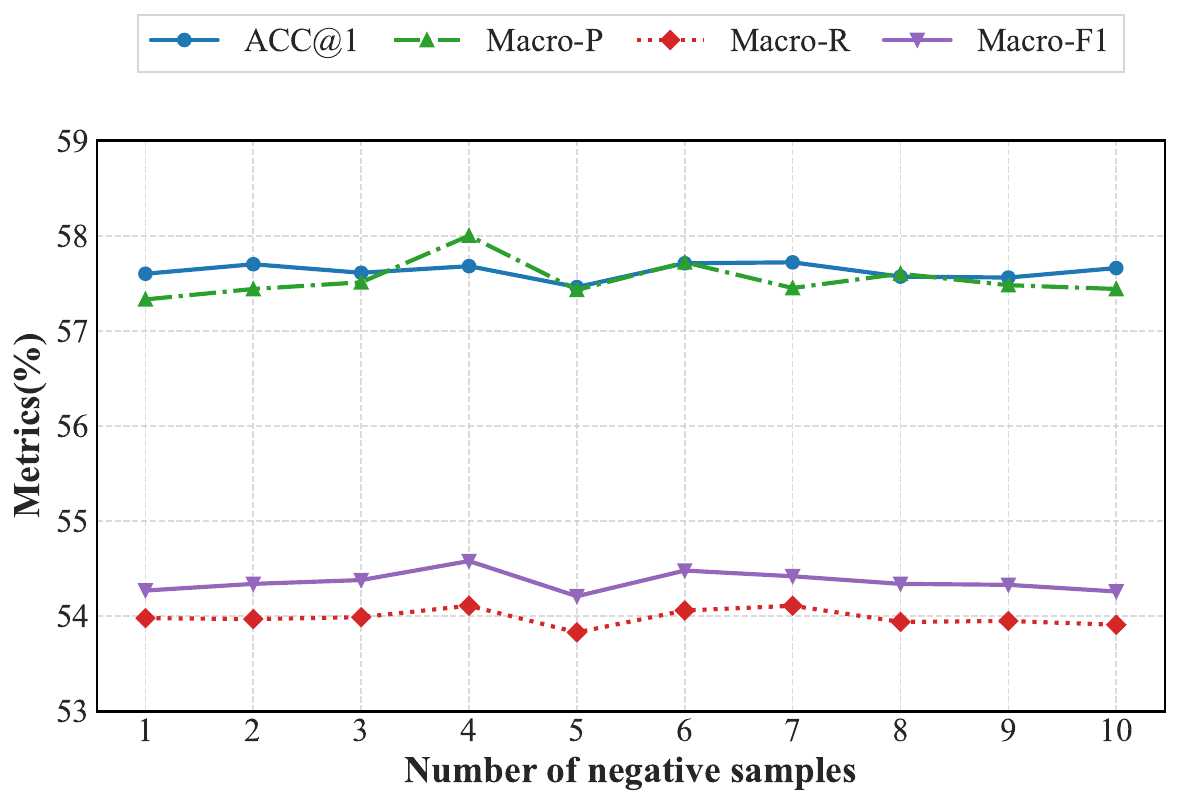}

    \small (d) Foursquare-TKY 1400
\end{minipage}

\medskip

\begin{minipage}{0.48\linewidth}
    \centering
    \includegraphics[width=\linewidth]
    {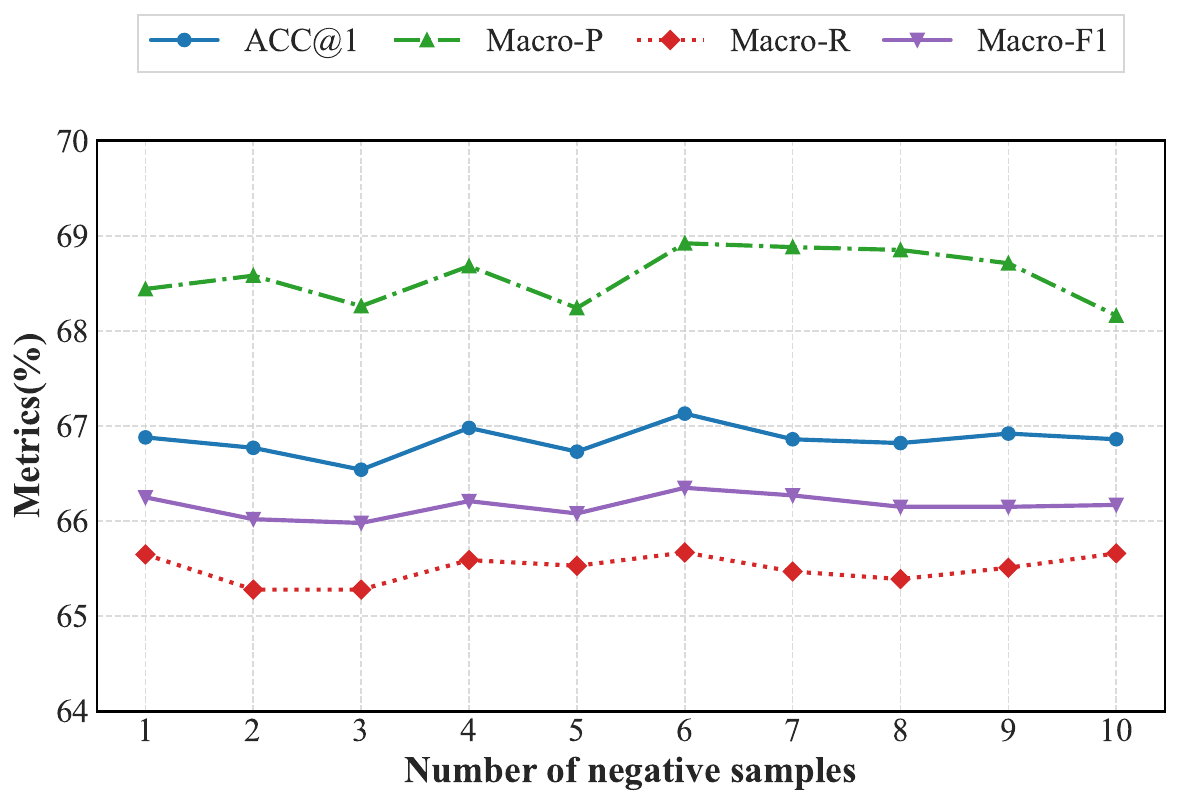}

    \small (e) Foursquare-JKT 700
\end{minipage}
\hfill
\begin{minipage}{0.48\linewidth}
    \centering
    \includegraphics[width=\linewidth]
    {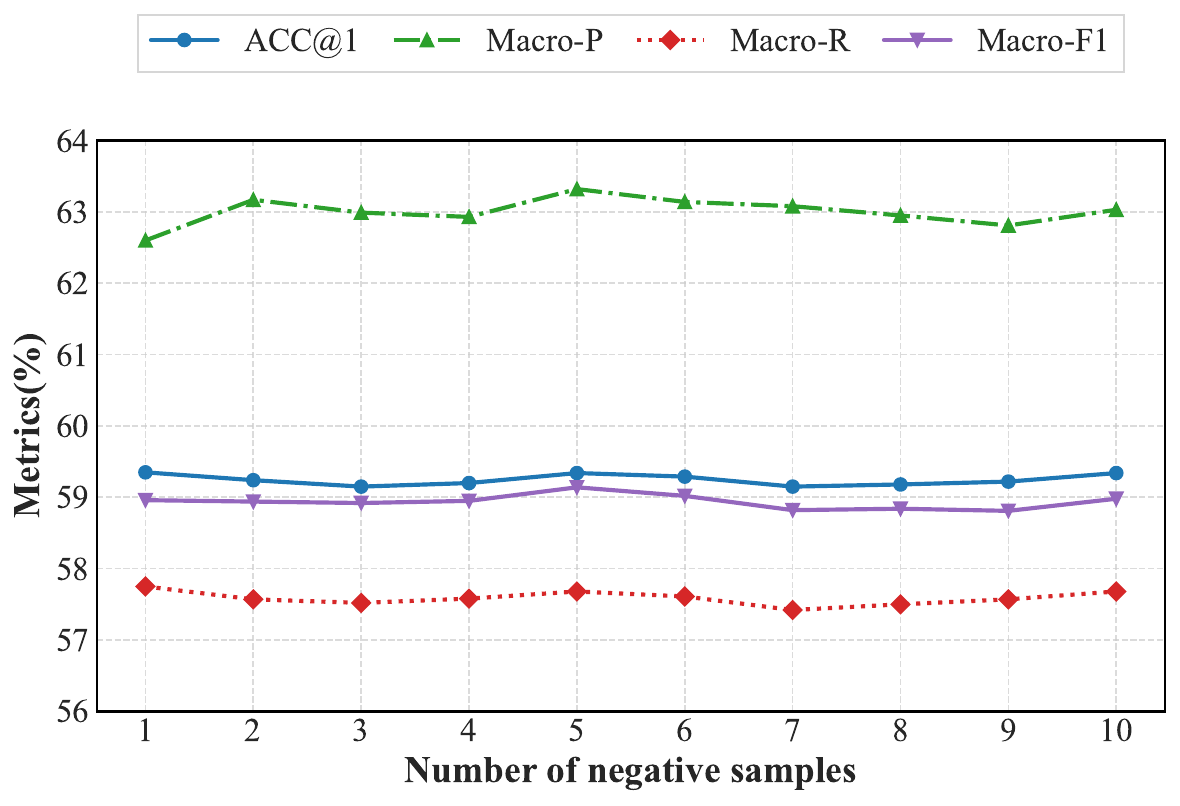}

    \small (f) Foursquare-JKT 1400
\end{minipage}

\captionof{figure}{Sensitivity analysis of the number of negative samples.}
\label{fig:negative_samples}

\end{center}

Figure~\ref{fig:negative_samples} shows the influence of the number of
negative samples, which is varied from 1 to 10. Increasing the number of
negative samples does not lead to a consistent improvement. Instead, the
metrics fluctuate within a relatively small range after a moderate number of
negative samples is used. More negative samples provide additional constraints
for knowledge graph learning but also increase the training cost. Based on
the overall results, 4 negative samples are used in our experiments as a
practical balance between model performance and computational efficiency.

\section{Conclusion}
\label{sec:conclusion}

This paper proposes MakeTUL, a Multi-Relational Knowledge Graph Enhanced Embedding framework for TUL. MakeTUL constructs a multi-relational knowledge graph to model visit-time, POI-category, and transfer relations, and learns POI and multimodal representations under these relational constraints. A trajectory prior structure extraction module further captures high-order POI co-occurrence patterns across trajectories, while the trajectory sequence learning module integrates structural, temporal, semantic, and transfer information to obtain trajectory representations. The global trajectory representation and sequence representation are finally combined through a dual-branch classification layer for trajectory owner prediction. Experiments on Foursquare-NYC, Foursquare-TKY, and Foursquare-JKT show that MakeTUL consistently outperforms the compared baselines across different user scales and evaluation metrics. The ablation and relation analysis further confirm the contribution of the main components and the complementary roles of different relation types. These results demonstrate that jointly modeling multi-relational semantics, trajectory structure, and sequential mobility information provides an effective approach to trajectory representation learning for TUL.

% \clearpage %%Remove this from your manuscript

\bibliographystyle{cas-model2-names}

\bibliography{refs}



\end{document}